\documentclass[make,article,submit,pdftex,moreauthors]{mdpi} 

\firstpage{1} 
\pubvolume{1}
\issuenum{1}
\articlenumber{0}
\pubyear{2024}
\copyrightyear{2024}
\datereceived{ } 
\daterevised{ } % Comment out if no revised date
\dateaccepted{ } 
\datepublished{ } 
\hreflink{https://doi.org/} % If needed use \linebreak
\usepackage{subcaption}
\usepackage{todonotes}
\usepackage{comment}
\usepackage{bm}
\Title{Parameter-efficient fine-tuning of large pretrained models for instance segmentation tasks}

\TitleCitation{Parameter-efficient fine-tuning of large pretrained models for instance segmentation tasks}

\Author{Nermeen  Abou Baker $^{1,\dagger,\ddagger}$\orcidA{}, David Rohrschneider $^{1,\dagger,\ddagger}$\orcidB{} and Uwe Handmann $^{1,\dagger}$\orcidC{}*}

\AuthorNames{Nermeen  Abou Baker, David Rohrschneider and Uwe Handmann} 

\AuthorCitation{Abou Baker, N.; Rohrschneider, D.; Handmann, U.}
\address{%
$^{1}$ \quad Ruhr West University of applied sciences}
\corres{Correspondence: Nermeen.Baker@hs-ruhrwest.de}%; Tel.: (optional; include country code; if there are multiple corresponding authors, add author initials) +xx-xxxx-xxx-xxxx (F.L.)}

\firstnote{Lützowstraße 5, 46236 Bottrop, Germany: Ruhr West University of Applied Sciences}  % Current address should not be the same as any items in the Affiliation section.
\secondnote{These authors contributed equally to this work.}

\abstract{Research and applications in artificial intelligence have recently shifted with the rise of large pretrained models, which deliver state-of-the-art results across numerous tasks. However, the substantial increase in parameters introduces a need for parameter-efficient training strategies. Despite significant advancements, limited research has explored parameter-efficient finetuning (PEFT) methods in the context of transformer-based models for instance segmentation. Addressing this gap, this study investigates the effectiveness of PEFT methods, specifically adapters and Low-Rank Adaptation (LoRA), applied to two models across four benchmark datasets. Integrating sequentially arranged adapter modules and applying LoRA to deformable attention—explored here for the first time—achieves competitive performance while finetuning only about $1-6\%$ of model parameters, a marked improvement over the $40-55\%$ required in traditional finetuning. Key findings indicate that using $2-3$ adapters per transformer block offers an optimal balance of performance and efficiency. Furthermore, LoRA, when applied to deformable attention, exhibits strong parameter efficiency and, in certain cases, surpasses adapter configurations. The results show that the impact of PEFT techniques varies based on dataset complexity and model architecture, underscoring the importance of context-specific tuning. Overall, this work demonstrates the potential of PEFT to enable scalable, customizable, and computationally efficient transfer learning for instance segmentation tasks.}

\begin{document}
%============== Introduction ==========================%
\section{Introduction}
\label{sec:intro}
Large pretrained models have gained increasing attention due to their ability to generalize across various tasks. Models such as the Segment Anything Model (SAM) \cite{kirillov2023segany} demonstrate this versatility by handling new tasks with zero-shot predictions. While earlier models, such as models based on convolutional networks, have proven effective in specific domains, their ability to generalize to new tasks remains limited  \cite{10.1007/978-3-031-43085-5_14}.\\
Traditional finetuning techniques involve adjusting the entire network or a subset of it to transfer knowledge to downstream tasks. However, this process often requires copying and updating the model’s weights for each task, resulting in high computational and memory costs. In addition, such methods can lead to catastrophic forgetting, where the model loses previously acquired knowledge when finetuned for new tasks. Despite their strong starting point, large pretrained models struggle to maintain their ability to predict out-of-distribution tasks. As the size of pretrained models continues to grow, the computational cost of finetuning for task-specific purposes increases significantly. Moreover, the risk of overfitting to target datasets further complicates the process.\\
To address these challenges, the emerging field of PEFT has introduced techniques that significantly reduce the number of trainable parameters while maintaining or approaching the performance levels of full finetuning. By modifying a small subset of parameters, PEFT not only reduces computational requirements but also mitigates the risks associated with overfitting.\\
Existing PEFT methods offer significant computational advantages, but face limitations that restrict their wider use in Computer Vision (CV) tasks. First, many techniques originally developed for Natural Language Processing (NLP) are not optimized for the unique challenges of vision tasks, such as instance segmentation, which requires precise pixel-level predictions and the ability to differentiate between object instances. Second, while these methods have succeeded in simpler vision tasks, their effectiveness in more complex scenarios such as instance segmentation and optimal use for models based on multi-scale deformable attention remains under-explored.\\\\
The motivation for this work stems from a notable gap in research regarding the application of PEFT techniques to contemporary DEtection TRansformer (DETR) models, particularly for instance segmentation tasks. Specifically, this study is motivated by two key areas of interest:
\begin{enumerate}
    \item The lack of prior literature addressing the integration of adapters and LoRA within deformable attention modules.
    \item Adapter modules usually yield limited scalability, as the bottleneck dimension is the primary adjustable parameter. Instead, a linearly scalable module can adapt more flexibly and efficiently to different model sizes and demands.
\end{enumerate}
To address these problems, as well as the limited exploration of PEFT techniques within the context of instance segmentation, this paper has the following contributions:
\begin{enumerate}
    \item First use of LoRA in deformable attention: A LoRA approach is introduced for multi-scale deformable attention, achieving parameter efficiency by only updating low-rank matrices in the attention layers.
    \item Sequential adapter integration for scalable finetuning: This study presents the first implementation of sequentially arranged adapter modules, particularly for two large pretrained instance segmentation models.
\end{enumerate}
The study aims to open new avenues of research by efficiently adapting large pretrained models to various vision applications. Additionally, the source code has been published to provide a valuable resource upon which the research community can build. 

\section{Finetuning techniques for large pretrained models in computer vision}
PEFT has proven highly effective in NLP, where models such as BERT and GPT are extensively finetuned for tasks such as sentiment analysis and machine translation \cite{QIU2024200308}. As large pretrained Vision Transformer (ViT) models gain importance in CV, there is a growing need for computationally efficient finetuning methods in this domain as well. PEFT techniques such as adapters, LoRA, prefix tuning and prompt tuning have shown great promise in reducing the number of trainable parameters, enabling a wide range of applications in NLP \cite{10.1145/3643787.3648039, NEURIPS2023_1feb8787, Zhang2023LLaMAAdapterEF, cooper-stickland-etal-2021-multilingual, bapna-firat-2019-simple} without the high cost associated with traditional finetuning. The main difference between these methods lies in their approach to parameter reduction and its impact on the model structure.

%========================== Adapters ==========================%
\subsection{Adapters}
Adapters are one of the earlier approaches to PEFT. They are small, trainable networks (adapter layers) between the layers of a pretrained model. Adapters typically work by adding a non-linear transformation that helps the model learn the task-specific features without changing the parameters of the main model. This reduces the number of trainable parameters required while mitigating the computational and memory costs associated with extensive model updates. Although this approach minimizes the adjustments to the main model parameters, it can introduce some complexity into the model architecture due to the inclusion of new layers. Moreover, these extra-added components to the model structure may increase inference time. Further details of the proposed adapter location and its architecture are described in section \ref{sec:models-and-arch:subsec:adapters-arch}.

%========================== LoRA ==========================%
\subsection{Low-Rank Adaptation (LoRA)}
LoRA is a PEFT technique designed to optimize large pretrained models by integrating trainable low-rank matrices into attention layers. This method approximates weight updates using these low-rank matrices, thereby reducing the number of parameters required for finetuning. During this process, the core parameters of the model remain largely frozen, which minimizes memory usage and computational cost while speeding up adaptation \cite{Hulora}. \\
LoRA's approach ensures that the architecture of the model remains efficient during inference, which is particularly effective for both NLP and vision tasks. By focusing on a lower-dimensional space for parameter tuning, LoRA addresses the challenges of finetuning large pretrained models, which often contain hundreds of millions of parameters. This technique aims to balance parameter efficiency and performance, making it a feasible solution for managing large-scale models in diverse applications.

%=================== Mathematical framework and efficiency of LoRA ==========================%
\subsubsection{Mathematical framework and efficiency of LoRA}
Given a pretrained model with a weight matrix \( W \in \mathbb{R}^{d \times k} \), the conventional finetuning approach updates \( W \) to \( W + \Delta W \). LoRA, instead, represents the update \( \Delta W \) as a product of two smaller matrices \( A \) and \( B \):
\begin{equation}
\Delta W = B \times A
\end{equation}
where \( B \in \mathbb{R}^{d \times r} \), \( A \in \mathbb{R}^{r \times k} \), and \( r \) is the rank of the decomposition, typically \( r \ll \min(d,k) \).
LoRA achieves a minimum amount of trainable parameters and computational complexity through this low-rank decomposition. The main are:
\begin{itemize}
    \item Parameter efficiency: Traditional finetuning requires updating \( d \times k \) parameters in \( \Delta W \). LoRA reduces this to \( r \times (d + k) \) parameters  (the sum of elements in the \( A \) and \( B \)) matrices, significantly reducing memory requirements.
    \item Training efficiency: While the original model computes \( W \times \text{input} \) with \( d \times k \) operations, LoRA adds only \( r \times (d + k) \) operations for \( (B \times A) \times \text{input} \). This means that during training, the original pretrained weights \( W \) are frozen, and only the low-rank matrices \( A \) and \( B \) are trained. This approach preserves the pretrained knowledge while adapting to new tasks.
    \item Inference optimization: During inference, LoRA computes the effective weights as:
    \begin{equation}
    W_{\text{LoRA}} = W + B \times A
    \end{equation}
    This allows the original weights \( W \) to remain unchanged, with only the smaller matrices \( A \) and \( B \)  being updated and stored.
    \item Empirical effectiveness: Studies have shown that LoRA can achieve performance comparable to full finetuning on various tasks, as detailed in section \ref{sec:related}.
\end{itemize}

%=================== LoRA and Hyperparameters ==========================%
\subsubsection{LoRA in transformer models and hyperparameters}
LoRA is typically applied to the attention layers in transformer-based models. Specifically, it modifies the query and value projection matrices in multi-head attention mechanisms.\\
LoRA has two main hyperparameters: Rank \( r \), which controls the trade-off between parameter efficiency and model capacity by adjusting the size of the decomposed weight matrices, and a scaling factor  \( \alpha \), which adjusts the magnitude of the LoRA update.

%=================== Prefix tuning ==========================%
\subsection{Prefix tuning}
It is another method that has been studied mainly in NLP. It involves freezing the parameters of the language model while optimizing a small, continuous, task-specific vector, known as the prefix. Inspired by prompting, prefix tuning allows subsequent tokens to attend to this prefix as if it were “virtual tokens” \cite{li-liang-2021-prefix}. However, prefix tuning performs poorly in many NLP tasks due to its instability during training and its reliance on flawed evaluation protocols \cite{chen-etal-2022-revisiting}.  As discussed in the original LoRA paper \cite{Hulora}, increasing the number of special tokens beyond a certain limit often leads to diminishing returns or performance drops in prefix tuning. In contrast, LoRA scales better and consistently matches or exceeds baseline finetuning, even in large models such as GPT-$3$, making it a more reliable approach for NLP tasks.
%Although it is effective in NLP, the sequential nature of prefix tuning makes it less suitable for vision tasks, where spatial complexity is a critical concern.
%=================== Prompt tuning ==========================%
\subsection{Prompt tuning}
Prompt tuning is a PEFT technique that introduces trainable \textit{soft prompts} into the input sequence of a frozen pretrained language model \cite{lester-etal-2021-power}. This approach simplifies adaptation by learning these soft prompts end-to-end, which can efficiently encapsulate task-specific signals and carry them throughout the model.\\
While prompt tuning performs well in LLMs for NLP, it has limitations in CV applications. The main drawback is that it relies on text processing, which is not directly applicable to visual data. Additionally, the performance of prompt tuning is constrained by the input capacity of the model, which limits the amount of task-specific information it can encode. As a result, this method is less effective for tasks that require detailed visual feature extraction or fine-grained adaptation, where specialized architectures or other finetuning methods may be more appropriate.\\ \\
Due to the above characteristics, this study explores the application of adapters and LoRA for instance segmentation tasks, and it focuses on their customization for large pretrained models. This research evaluates the effectiveness of these parameter-efficient techniques in enhancing the performance of instance segmentation applications.
%=================== Related work ==========================%
\section{Related work}\label{sec:related}
\subsection{Adapters in vision tasks}
\label{sec:related-adapters}
The idea of using adapters together with transformers was first introduced in NLP research with a compact and scalable architecture \cite{pmlr-v97-houlsby19a}. It was designed for multi-tasking and continual learning by using a single BERT model, that is shared among task-specific parameters. The \textit{Vision Transformer Adapter} then transferred this idea to CV, allowing a plain ViT \cite{ViT} to be adapted to multiple dense prediction tasks \cite{Chen2022VisionTA}. The authors found that the plain variant reached its limits when transferred to tasks such as object detection or instance segmentation, and so designed a parallel architecture to extract, augment, and inject task-specific information along the entire ViT.\\
Recent advances in this area include SAM, which has recently received increased attention for providing accurate segmentation masks based on different types of user prompts \cite{kirillov2023segany}. Previous attempts to finetune SAM have been investigated, as shown by Abou Baker et al. \cite{ESANN-BakHan2023}. To leverage the base knowledge of the model learned during pretraining, this approach involved finetuning only the decoder, resulting in significant improvements that outperformed the state-of-the-art (SOTA) of multiple waste datasets. Further, Chen et al. \cite{Chen2023SAMFT} additionally implemented adapter modules to finetune SAM for the detection of camouflaged objects, shadows, and polyps in under-represented scenes. The authors demonstrate the effectiveness of these modules by outperforming the segmentation capabilities of the plain SAM on $5$ different datasets.\\
In the medical image segmentation domain, the Medical SAM Adapter (Med-SA)\cite{wu2023medical} added bottleneck adapters to both the encoder and decoder components. This method achieved superior performance over SOTA medical image segmentation techniques while updating only $2\%$ of the model parameters during finetuning.\\
In multi-modal large pretrained models, CLIP, which originally combined visual and text embeddings for visual classification tasks \cite{radford2021learning}, was further enhanced with adapters in the work of Gao et al. \cite{Gao2021CLIPAdapterBV}. The authors added a bottleneck adapter after both the language and vision backbone and combined the result with the initial embeddings via a skip connection. Consequently, the performance could be improved over $11$ datasets.
\subsection{LoRA in vision tasks}
\label{sec:related-lora}
Large pretrained models, trained using self-\textbf{DI}stillation with \textbf{NO} labels (also called DINO) \cite{Caron2021EmergingPI} and DINOv$2$ \cite{oquab2024dinov} have shown exceptional performance in many vision tasks, but certain limitations have constrained their application in medical and surgical contexts. However, significant progress has been made in several areas, addressing these challenges through LoRA techniques. For example, Zhang et al. \cite{zhang2024learningadaptfoundationmodel} integrate LoRA layers into the DINOv$2$ model to improve diagnostic accuracy in capsule endoscopy. By freezing the parameters of the core model and selectively applying LoRA to the query and value projection layers in the transformer blocks, this method addresses the challenges posed by limited medical datasets and improves the adaptability for image classification. Evaluations on the Kvasir-Capsule and Kvasir-v$2$ datasets result in impressive results, with the adapted model achieving accuracies of $97.75\%$ and $98.81\%$, respectively, outperforming several SOTA Convolutional Neural Network (CNN) and ViT models. Additionally, the LoRA-based adaptation reduced training time and memory requirements compared to full finetuning, while maintaining the strong visual representation capabilities of the DINOv$2$ encoder. These results emphasize the efficiency of LoRA in adapting large pretrained models for specialized medical tasks such as capsule endoscopy diagnosis.\\
Similarly, Cui et al. \cite{Cui2024SurgicalDINOAL} focused on depth estimation in robotic surgery, which is critical for tasks such as $3$D reconstruction, surgical navigation, and augmented reality visualization. The study introduced LoRA of the DINOv$2$ model, called Surgical-DINO, specifically designed for depth estimation in endoscopic surgery. By integrating LoRA layers and keeping the DINO image encoder frozen, this approach allows for effective adaptation to surgery-specific domain knowledge without extensive finetuning. The Surgical-DINO model was evaluated on the MICCAI SCARED dataset and demonstrated superior performance compared to SOTA models in endoscopic depth estimation. The study shows that zero-shot prediction is insufficient for this case and that LoRA adaptation is crucial for achieving high performance. Additionally, the study refers to a gap in current research, which focuses on segmentation and detection tasks, rather than pixel-wise regression tasks such as depth estimation. This work demonstrates how adapting large pretrained models with LoRA can significantly improve their applicability to specialized medical tasks, providing a promising direction for future research.\\
Another study investigated the application of LoRA to unsupervised domain adaptation for semantic segmentation tasks, addressing the challenges posed by large transformer models and their computational requirements \cite{10.3103/S1060992X2306005X}. The study used LoRA to transfer models from synthetic datasets (GTA5) to real-world datasets (Cityscapes) and focused on improving training stability and efficiency. The authors integrated LoRA into the Swin transformer and TransDA framework and showed that LoRA effectively stabilized the self-training process. This approach achieved performance comparable to the exponential moving average (EMA) mechanism while reducing training time and memory usage by $11\%$. The study demonstrated the potential of LoRA as a computationally efficient alternative for domain adaptation and illustrated its effectiveness in improving semantic segmentation performance with reduced resource requirements.\\
Generalized LoRA (GLoRA) \cite{chavan2023oneforallgeneralizedloraparameterefficient} is an advanced method for PEFT that significantly improves performance on the VTAB-$1$K benchmark. GLoRA extends LoRA by adding a generalized prompt module to optimize both model weights and intermediate activations, providing greater flexibility and efficiency for a variety of tasks. The study shows that GLoRA outperforms existing methods in the VTAB-$1$K benchmark by up to $2.9\%$, achieving superior accuracy with fewer parameters and reduced computational requirements. GLoRA can maintain high generalization capabilities across $14$ out of $19$ datasets proving its effectiveness in bridging the gap between parameter efficiency and model performance.\\
To address the challenge of efficiently finetuning large models for image generation, SuperLoRA is a novel framework that extends LoRA for finetuning large models \cite{chen2024superloraparameterefficientunifiedadaptation}. SuperLoRA extends traditional LoRA by including techniques such as grouping, folding, shuffling, projecting, and tensor factoring, which significantly improves flexibility and performance, especially in scenarios with extremely limited parameters. The work is evaluated through transfer learning tasks, including image classification and image generation, and demonstrates superior parameter efficiency compared to existing LoRA variants. SuperLoRA achieves up to $10$-fold reduction in the number of parameters while maintaining or improving performance, proving highly effective in low-parameter scenarios.\\
Building on this, a recent study applied LoRA to finetune a stable diffusion model for generating synthetic images of defects (e.g., scratches, cracks, and pits) in the NEU-seg dataset \cite{s24186016}. To overcome data scarcity and class imbalance, synthetic images were used to augment the training data for segmentation models (DeepLabV3+ and FPN). This method led to significant improvements in segmentation performance, with the mean Intersection-over-Union (mIoU) increasing by about $6\%$. The key contribution is the use of LoRA to efficiently adapt stable diffusion for high-quality synthetic image generation, which improved the robustness of defect segmentation.\\
The previous related work shows that adapters and LoRA techniques are effective in various domains such as NLP and certain CV tasks such as image classification and semantic segmentation, as summarized in Table \ref{related-work}. However, their application to more specialized and complex CV problems, particularly instance segmentation, remains under-explored.\\
Instance segmentation presents a unique challenge because it requires not only the identification of semantic classes at the pixel level but also the grouping of these pixels into distinct object instances. This task involves separately predicting both the mask and the category for each object instance, adding a layer of complexity beyond what existing adaptation methods in CV have primarily supported.\\
\def\arraystretch{1.2}
\begin{table}[h!]
\caption{Summary of related work and their classification into the topics \textit{Adapters in CV} and \textit{LoRA in CV}.}
\label{related-work}
\centering
    \resizebox{\textwidth}{!}{
    \begin{tabular}{
        |p{0.8\textwidth}
        |>{\centering}p{0.1\textwidth}
        |>{\centering\arraybackslash}p{0.08\textwidth}|
    } \hline 
    \textit{Publication} & \textit{Adapters in CV} & \textit{LoRA in CV} \\\hline

    Vision Transformer Adapter for Dense Predictions \cite{Chen2022VisionTA} & \checkmark & \\\hline
    
    SAM-Adapter: Adapting Segment Anything in Underperformed Scenes \cite{Chen2023SAMFT} & \checkmark & \\\hline

    Medical SAM adapter: Adapting segment anything model for medical image segmentation \cite{wu2023medical} & \checkmark & \\\hline

    Clip-adapter: Better vision-language models with feature adapters \cite{Gao2021CLIPAdapterBV} & \checkmark & \\\hline

    Learning to Adapt Foundation Model DINOv2 for Capsule Endoscopy Diagnosis \cite{zhang2024learningadaptfoundationmodel} & & \checkmark \\\hline

    Surgical-DINO: adapter learning of foundation models for depth estimation in endoscopic surgery \cite{Cui2024SurgicalDINOAL} & & \checkmark \\\hline
    
    Low Rank Adaptation for Stable Domain Adaptation of Vision Transformers \cite{10.3103/S1060992X2306005X} & & \checkmark \\\hline
    
    Latent Diffusion Models to Enhance the Performance of Visual Defect Segmentation Networks in Steel Surface Inspection \cite{s24186016} & & \checkmark \\\hline

    This work & \checkmark & \checkmark \\\hline
    \end{tabular}}
\end{table}\\
To the best of our knowledge, this work pioneers the application of adapters and LoRA to instance segmentation using large pretrained models. By addressing this under-explored area, this research aims to extend the capabilities of adaptation methods to handle complex visual tasks and contribute to the advancement of efficient, adaptable CV models for instance segmentation.
%=================== Models and architecture ==========================%
\section{Models and architecture}
\label{sec:models-and-arch}
%======================= large pretrained models =======================%
\subsection{Large pretrained models}
\label{sec:models-and-arch:subsec:large-pretrained-models}
This study examines two large pretrained models specialized for instance/panoptic segmentation:
Segment Everything Everywhere Model (SEEM) \cite{zou2023segment} and DETR with Improved Denoising Anchor Boxes (Mask DINO) \cite{li2022mask}. SEEM is a multi-modal segmentation model that accepts prompts like clicks, boxes, polygons, scribbles, text, and referring regions from another image to control the focus of the detection during inference. Inspired by the architecture of CLIP \cite{radford2021learning}, it also incorporates a text encoder to enable open-set segmentation capabilities and thus can perform so-called "zero-shot inference" on unseen datasets. SEEM, with approximately $340$ million parameters, provides a robust framework for various types of input data.\\
The interactive models are highly effective in helping with image labeling, which remains a major challenge in CV. In the work of Huang et al. \cite{9500146}, they propose a novel attention mechanism that focuses primarily on the scribbles to enhance the contrast between background and foreground objects. This approach reduces the need for dense annotations, increases the efficiency of automatic labeling, and achieves high-quality segmentation with minimal user input.
In a more recent work by Meta, SAM $2$ extends its predecessor by allowing the segmentation of specific objects throughout videos with one or more interactive prompts on the first frame. It is designed to provide a real-time experience while achieving similar or better results than SAM \cite{ravi2024sam2}.\\
Mask DINO, with two versions provided by the authors: one based on the ResNet backbone (with $52.02$ million parameters), and another based on the Swin-L backbone (with $222.76$ million parameters), is a large pretrained model for object detection and segmentation. It adopts the same architecture as the plain DINO model but modifies the transformer decoder slightly to enable the output of instance-wise segmentation masks. This is achieved by adding a parallel branch dedicated to mask prediction alongside DINO's bounding box prediction branch. It outperforms all existing instance segmentation methods that are based on the ResNet$50$\cite{ResNet} and Swin-L backbone \cite{liu2021swin}, achieving $54.5$ Average Precision (AP) on the COCO dataset.\\
The methodology of this work has been applied to SEEM and Mask DINO, as both are based on the segmentation \textit{meta-architecture} proposed by the authors of MaskFormer \cite{cheng2021perpixel} and deformable DETR \cite{deformable-detr}, simplifying the process of code adjustments during implementation. Additionally, they offer a variety of backbone methods to choose from, including various sizes of FocalNet \cite{yang2022focal} or ViT \cite{ViT} for SEEM and ResNet$50$ \cite{ResNet} or Swin-L \cite{liu2021swin} for Mask DINO. Unlike SEEM, Mask DINO is not a multi-modal model, so the proposed techniques can be compared for both types of large pretrained models. As a replacement for the regular multi-headed self-attention layer, both models are implemented to support the use of multi-scale deformable attention. However, only the authors of Mask DINO made use of this replacement during the creation of the publicly available weights, which caused a minor architectural difference between the transformer blocks compared to SEEM.
%========================== Architecture =========================%
\subsection{Architecture}\label{sec:models-and-arch:subsec:architecture}
\begin{figure}[h!]
    \centering
    \includegraphics[width=\textwidth]{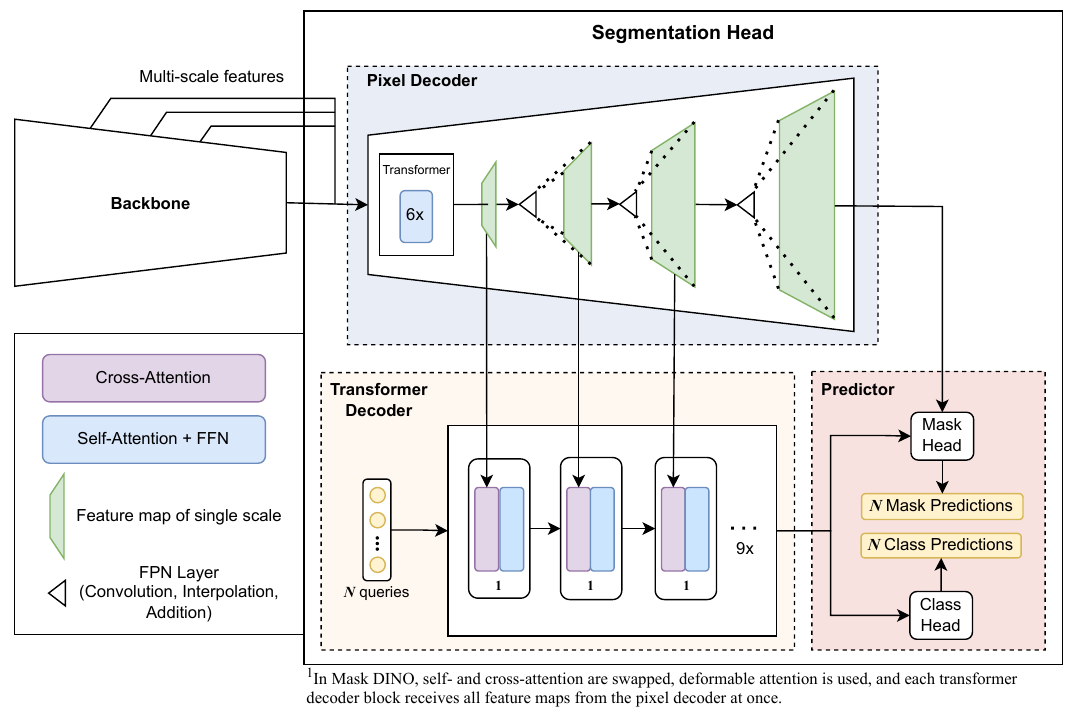}
    \caption{High-level view on the model architecture of Mask DINO and SEEM.}
    \label{fig:meta-arch}
\end{figure}

Figure~\ref{fig:meta-arch} provides a high-level overview of the previously mentioned \textit{meta-architecture}. The models consist of four components: The \textbf{backbone}, which produces multiple image embeddings of different sizes, \textbf{a pixel decoder} to upsample the embeddings into high-resolution feature maps gradually, a \textbf{transformer decoder} to generate $N$ embeddings for possible instance candidates, and a \textbf{predictor} to convert and filter these candidates to actual class and mask predictions. The pixel decoder is built upon a three-level Feature Pyramid Network (FPN) \cite{lin2017feature}, enhanced by a transformer encoder with \(6\) layers to capture the global image features, and efficiently extract high-resolution per-pixel features.\\
Each of the $9$ transformer decoder blocks consists of three layers: A cross-attention module with layer normalization (LN), which receives the image features produced by the pixel decoder; a self-attention module with LN; and a feed-forward network (FFN) with LN - based on the method proposed by Cheng et al. \cite{cheng2022maskedattention}. Inside the predictor, the output of the transformer decoder is combined with a learnable \textit{class embedding}, resulting in the class logits. In Mask DINO, a softmax function generates class probabilities. SEEM, as an open-set segmentation model, calculates the cosine similarity between these class logits and the embeddings of the textual input to derive the probabilities.
Mask predictions are obtained by feeding the same decoder output through a simple MLP, called \textit{mask embedding}, and linearly combining these outputs with the high-resolution feature map of the \textit{pixel decoder}. In Figure~\ref{fig:meta-arch}, these operations take place in the Mask and Class Head, respectively.
Throughout this work, the composition of all the modules after the backbone is called \textit{segmentation head}.
%============================ Adapters ===========================%
\subsection{Adapters}\label{sec:models-and-arch:subsec:adapters-arch}
The proposed adapter architecture is inspired by the NLP-based approach introduced by Houlsby et al. \cite{pmlr-v97-houlsby19a}. In their approach, a linear layer is used to project down the output of the intermediate transformer layer, an activation function is used to introduce non-linearity at the bottleneck, a linear layer is used to project up to the original input size, and a residual connection is established to preserve important information about the input signal. In the original paper, the adapters were placed after each feed-forward network of the BERT transformer layers, resulting in $2$ adapters per layer. Considering these aspects, the proposed method introduces three main architectural decisions to transfer and extend this idea to segmentation frameworks.
%=================== Adapter location ==========================%
\subsubsection{Adapter location}\label{sec:models-and-arch:subsec:adapters-arch:subsubsec:adapter-location}
\begin{figure}[h!]
    \centering
    \includegraphics[width=\textwidth]{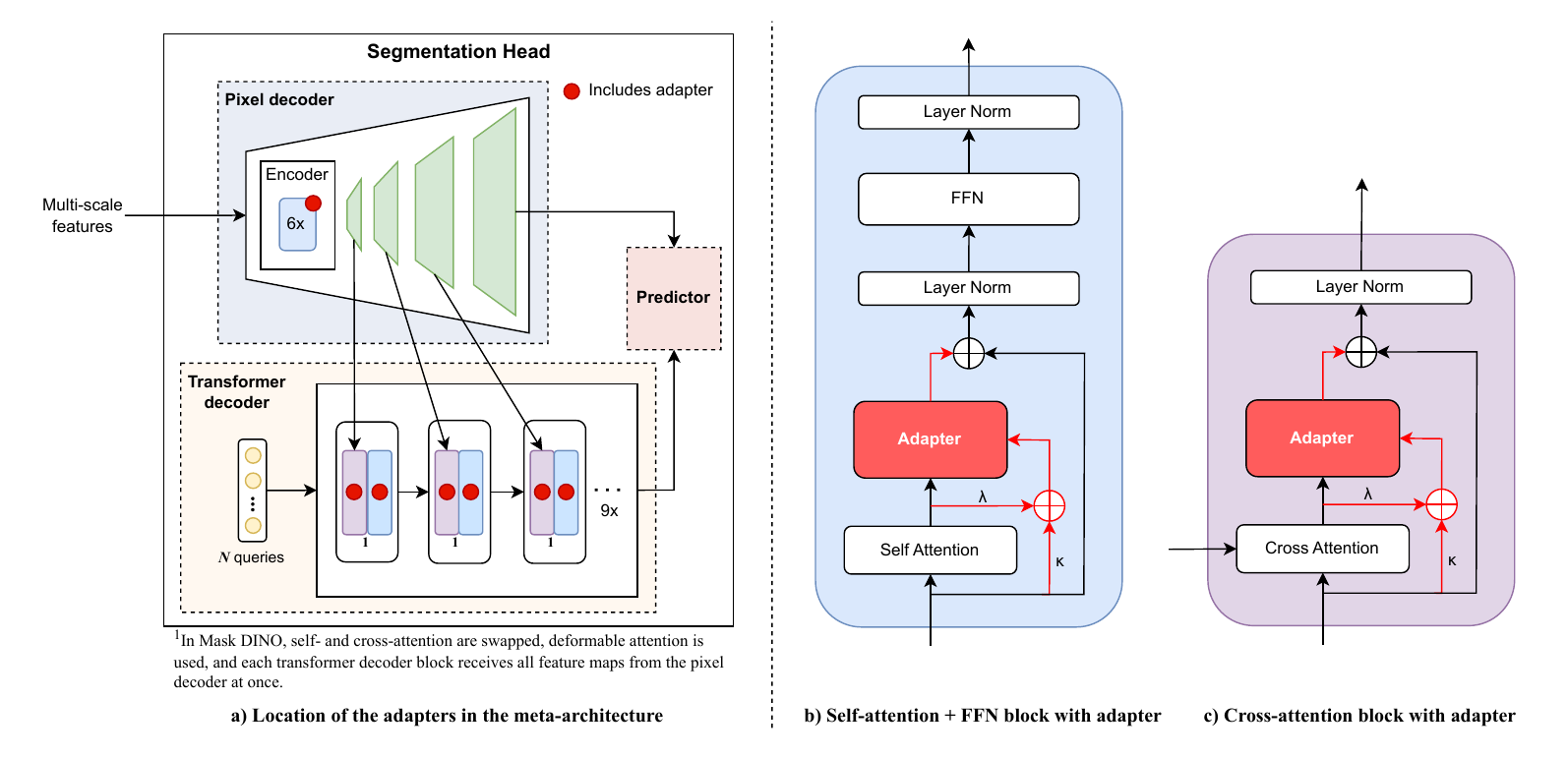}
    \caption{Location of adapters inside the meta-architecture (a) and detailed placement inside a self- or cross-attention block (b + c). Added components are colored red.}
    \label{fig:adapter-location}
\end{figure}
Since the main purpose of the adapters is to "translate" the knowledge of the pretrained transformer to a downstream task, the adapter modules have been placed directly after all cross- and self-attention layers of each transformer block (highlighted in red-colored blocks in Figure~\ref{fig:adapter-location}b) and c) below). This location ensures that the adapters receive the original attention maps of the network as a prior and are trained to apply the non-linear transformation. Consequently, the outputs can be seen as modified attention maps that contain additional task-specific information for the downstream tasks.
%=================== Adapter architecture ==========================%
\subsubsection{Adapter architecture} \label{sec:models-and-arch:subsec:adapters-arch:subsubsec:adapter-arch}
\begin{figure}[h!]
    \centering
    \includegraphics[width=0.95\textwidth]{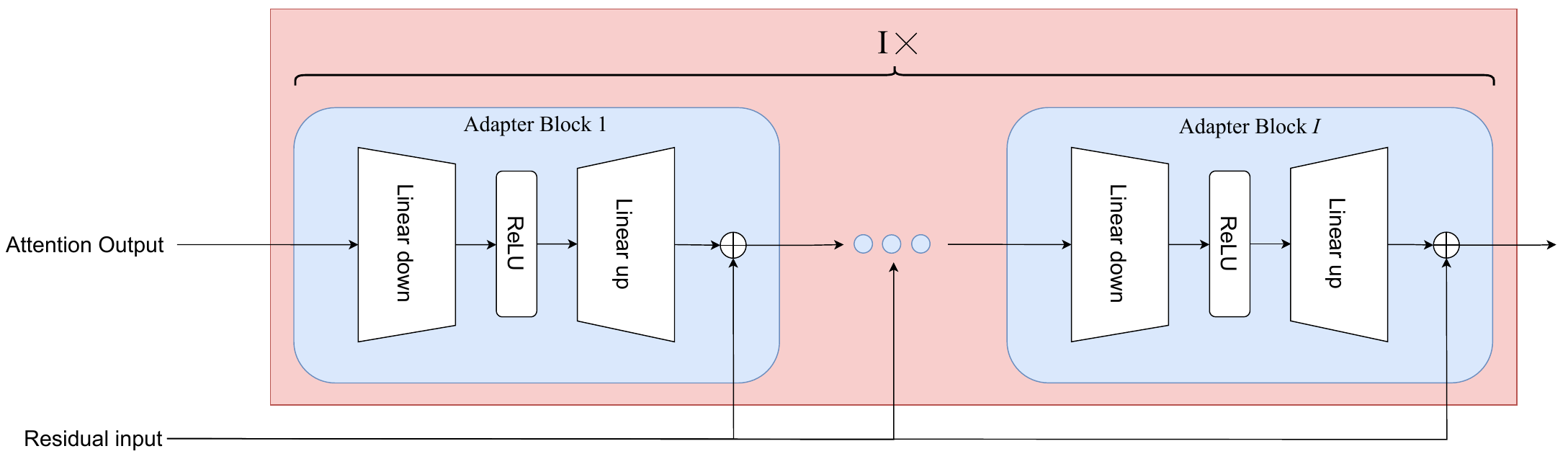}
    \caption{An adapter block consists of a simple feed-forward network and a residual connection (blue block). Inside one adapter (red frame), this block is sequentially repeated $I$ times. Each of the $I$ blocks receives the same residual input.}
    \label{fig:adapter-arch}
\end{figure}

The internal adapter architecture closely follows the original implementation mentioned earlier \cite{pmlr-v97-houlsby19a}, consisting of a linear down-projection, a ReLU activation function, a linear up-projection, and a residual input that is added ($\oplus$) to the adapter output, as shown in Figure~\ref{fig:adapter-arch} (The blue block). The bottleneck dimension is defined and calculated as follows:
\begin{equation}
    dim_{bottleneck} = dim_{input} * \frac{1}{k} \hfill ,
\end{equation}
where \(k\) is the downscale factor, a tunable hyperparameter shared by all adapter modules. Increasing \(k\) reduces the dimension of the bottleneck, effectively leading to a common trade-off between compressing the data to capture high-level relations and losing relevant information due to massive dimension reduction. Furthermore, since the attention prior is highly informative and may require a deeper neural network than just a single MLP to capture all relevant information, another hyperparameter $I$ defines the number of repetitions of the MLP within each adapter module. Fundamentally, all repetitions receive the same residual input to refer to the original signal throughout the sequential alignment.
%=================== Residual connection ==========================%
\subsubsection{Residual connection}\label{sec:models-and-arch:subsec:adapters-arch:subsubsec:adapter-residual}
In contrast to the original implementation, where the residual input serves as a skip connection to add up the information before the adapter module, it was observed that combining the attention output with the key vector of the attention input yields slightly better overall performance, as discussed later in section~\ref{sec:ablation}. This improvement occurs even though the models are designed to include this signal inherently and to add it to the output of the adapter in both cases (as shown in Figures~\ref{fig:adapter-location}b and \ref{fig:adapter-location}c above the adapter blocks). The interpretation of this behavior is that emphasizing the attention's key vector \textbf{after} the adapter modules forces the adapters to focus more on learning how to finetune the pretrained attention layers, rather than learning the relationships between the input features from scratch. Using the attention prior only as a residual input is referred to as the $\kappa$ configuration. In contrast, the residual use of the attention output is referred to as the $\lambda$ configuration. The combination of both, shown in Figures~\ref{fig:adapter-location}b and \ref{fig:adapter-location}c, is referred to as the $\kappa + \lambda$ configuration.
%============================ LoRA  layers ===========================%
\subsection{LoRA layers}\label{sec:models-and-arch:subsec:lora-layers}
\begin{figure}[h!]
    \centering
    \includegraphics[width=0.95\textwidth]{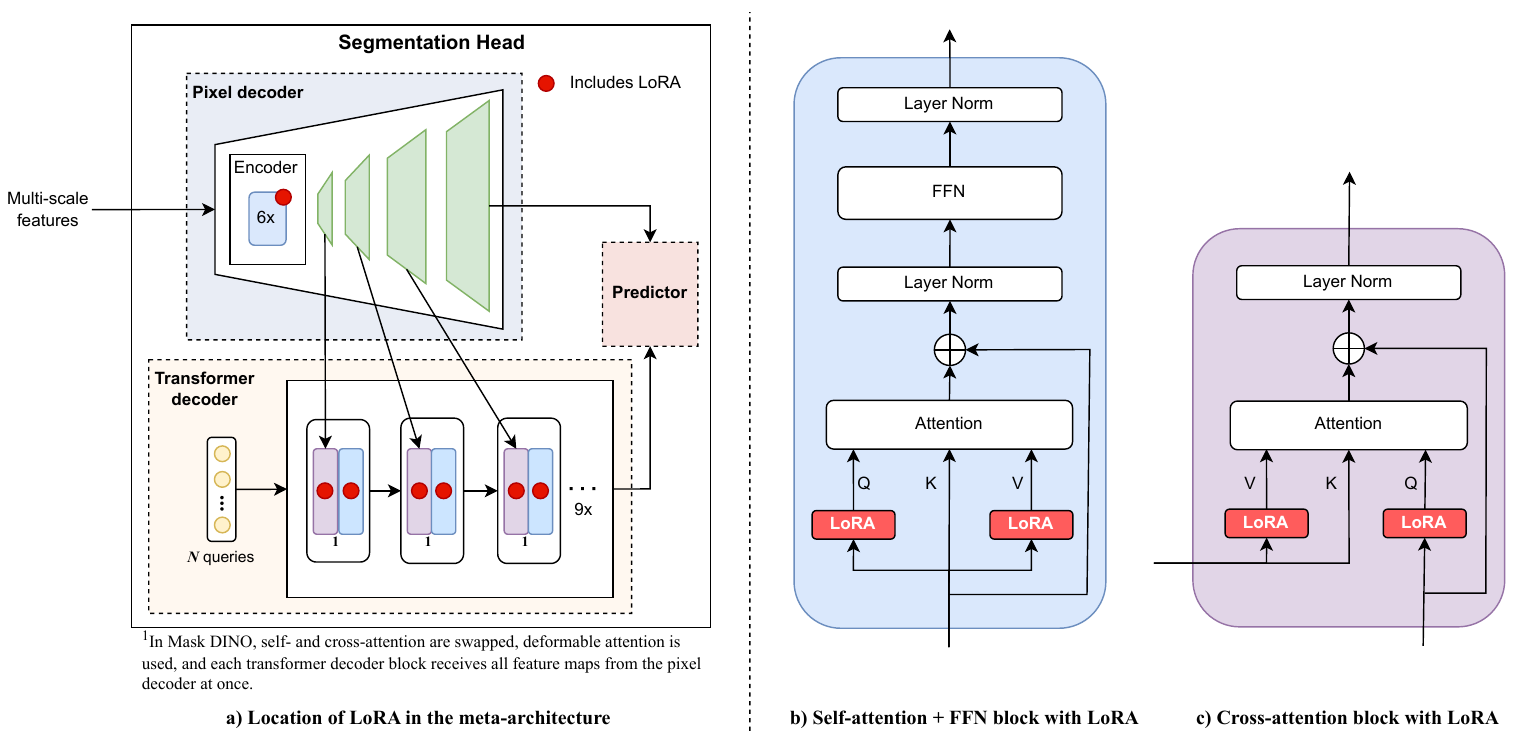}
    \caption{Location of LoRA inside the meta-architecture (a) and detailed placement inside a self- or cross-attention block (b + c). Added components are colored red.}
    \label{fig:lora-location}
\end{figure}

\begin{figure}[h!]
    \centering
    \includegraphics[width=\linewidth]{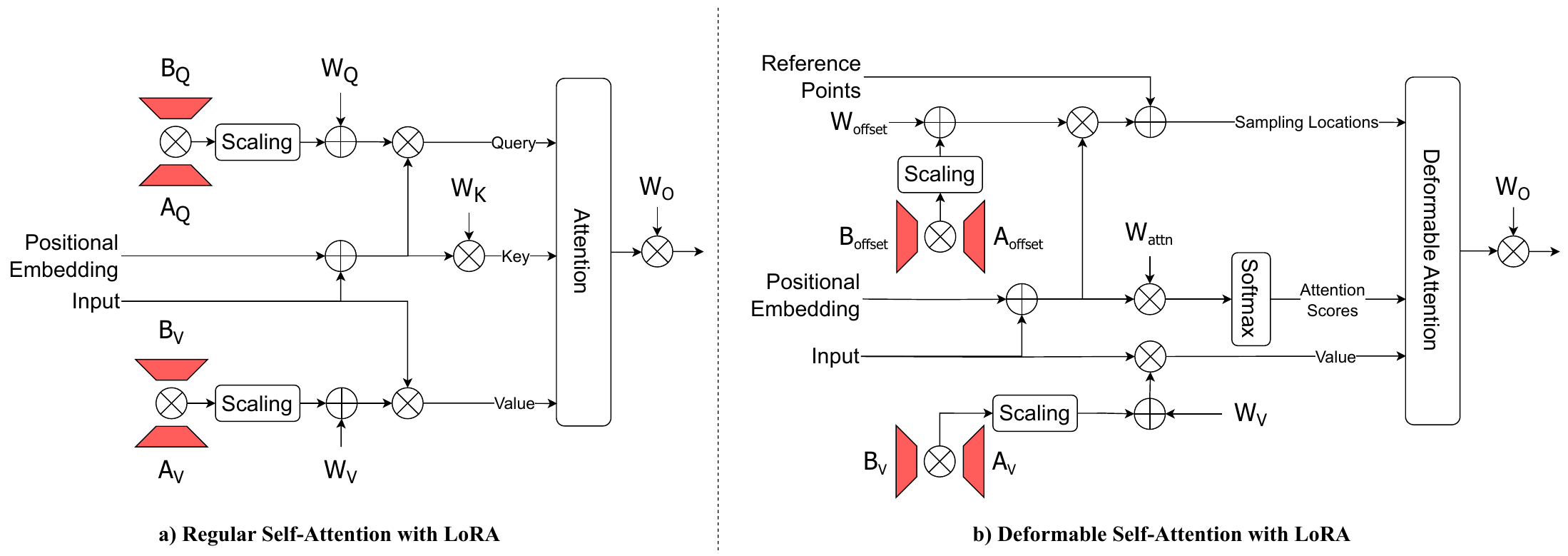}
    \caption{The LoRA application to regular and deformable attention mechanism.}
    \label{fig:lora-layers}
\end{figure}

Similar to the adapter modules, LoRA was applied to each of the \(6\) transformer encoder layers within the pixel decoder and to all $9$ blocks of the transformer decoder, as illustrated in Figure \ref{fig:lora-location}.\\
\\
According to the original LoRA implementation \cite{Hulora}, the input sequence is processed in parallel by the base weights (\(W_0\)) and the LoRA weights (\(BA\)), scaled by a factor \(\frac{\alpha}{r}\). \textbf{Then}, the results are added, leading to the following expression:
\begin{equation}
\label{eq:standard-lora}
    h = x W_{0}^T + \frac{\alpha}{r}x (BA)^T
\end{equation}
However, this calculation involves applying two matrix multiplications (one with the base weights and another with the LoRA weights) to the input sequence. Thus, the LoRA weights must also be kept separate during inference, potentially increasing the inference speed compared to the plain model. To keep the latency as low as possible, it was decided to use the "LoRA-Torch" Python library \cite{lin2023loratorch}. It efficiently bypasses the issue by \textbf{first} adding the scaled LoRA weights to the base weights and \textbf{then} applying a single matrix multiplication to the input sequence, leading to the following equation:
\begin{equation}
\label{eq:efficient-lora}
    h = x (W_0 + \frac{\alpha}{r}BA)^T
\end{equation}
For each task adaptation, the weight matrix \(B\) is initialized as a zero matrix, allowing the model to begin with the pretrained weights. As training progresses, task-specific weights are added incrementally, ensuring that the model can learn new tasks while preserving the knowledge encoded in the pretrained parameters.
During training, the re-composition of $A$ and $B$ is performed before each forward pass and then reversed by subtracting the scaled LoRA weights. This process isolates the contribution of LoRA-specific weights from the base weights, allowing the loss to be correctly back-propagated. For inference, the trained weights can be pre-computed and then added to the model's weights to generate a finetuned version of the model, minimizing the computational latency during inference.\\
Since the authors of Mask DINO replaced the regular self-attention in the pixel decoder and the cross-attention layers in the transformer decoder with the deformable attention mechanism, this work proposes a paradigm to apply LoRA to both variants while maintaining its functionality. This distinction and the previously mentioned weight addition are illustrated in Figure \ref{fig:lora-layers} and described below.
\subsubsection{Regular self- and cross-attention}
In the regular self- or cross-attention mechanism, \(Q\), \(K\) and \(V\) are calculated by a matrix multiplication between the input sequence and corresponding weight matrices \(W_Q\), \(W_K\) and \(W_V\). After the actual attention operation, the output is multiplied by a fourth set of weights (\(W_O\)) to apply a final linear transformation. The operation is expressed mathematically in Equation \ref{eq:regular-attention}. For simplicity, it represents the self-attention operation with a only single head:
\begin{equation}
\label{eq:regular-attention}
    \begin{aligned}
        &Q = xW_Q^T, K = xW_K^T, V = xW_V^T \\
        &Output = Attention(Q, K, V)W_O^T
    \end{aligned}
\end{equation}
Based on findings from the original LoRA paper \cite{Hulora}, applied to the weight matrices of the query (\(W_Q\)) and value (\(W_V\)) projections, leaving the key (\(W_K\)) and output (\(W_O\)) projections untouched. To achieve this, the equation \ref{eq:regular-attention} is extended as follows:
\begin{equation}
\label{eq:regular-attention-lora}
    \begin{aligned}
        &Q = x(W_Q + \frac{\alpha}{r}B_QA_Q)^T \\
        &K = xW_K^T \\
        &V = x(W_V + \frac{\alpha}{r}B_VA_V)^T \\
        &Output = Attention(Q, K, V)W_O^T,
    \end{aligned}
\end{equation}
where the decomposed LoRA matrices ($A_Q$, $B_Q$, $A_V$, $B_V$) are trained during finetuning.
In Figure \ref{fig:lora-layers}a), the trainable matrices are visualized in red. After re-composition, they are added (\(\oplus\)) to the base weights and the remainder follows the conventional attention mechanism. 
For cross-attention, LoRA was applied to the same weight matrices as in self-attention. The only difference is that the key and value matrices are derived from a different input sequence than the query matrix, allowing the model to attend to an external source of information.
\subsubsection{Deformable self- and cross-attention}
The regular attention mechanism involves calculating attention scores between all elements of the query and key sequences, resulting in quadratic complexity with respect to the sequence length. This issue becomes particularly concerning for images, where computational complexity scales quadratically for \textbf{both} height and width.\\
To mitigate this issue, the authors of Zhu et al. \cite{deformable-detr} proposed the deformable attention mechanism, which restricts each element to only attend to a fixed number of points in the sequence. This approach enables the computational feasibility of executing self-attention operations on large sequences, such as the flattened concatenation of image patches obtained from multiple scales of the backbone architecture.\\
To accomplish this, \(K\) static reference points (\(\hat{p}\)) are predefined as an evenly distributed grid within the spatial dimensions of the input image. Based on the query sequence, the model predicts offsets (\(\Delta p\)) and attention weights (\(\hat{A}\)) for each reference point. These offsets are added to the reference points to generate sampling locations (\(p\)).\\
Once the offsets are applied, a softmax function is used to normalize the attention weights (\(A\)). The corresponding points are sampled from the value sequence and weighted according to these normalized attention scores. To produce the final output, a linear transformation using another weight matrix \(W_O\) is applied to the result.\\
In the original implementation \cite{deformable-detr}, the offsets are calculated by multiplying the query sequence and a weight matrix \(W_{offset}\). Similarly, the attention weights are obtained with a separate weight matrix \(W_{attn}\), and the value projection remains the same as in the regular attention mechanism (\(W_V\)).
Equation \ref{eq:deformable-attention} expresses the deformable self-attention operation mathematically, simplified only to include a single head:
\begin{equation}
\label{eq:deformable-attention}
    \begin{aligned}
        &\Delta p = xW_{offset}^T, \hat{A} = xW_{attn}^T, V = xW_V^T \\
        &p = \hat{p} + \Delta p \\
        &A = softmax(\hat{A}) \\
        &Output = DeformableAttention(V, p, A)W_O^T
    \end{aligned}
\end{equation}
Deformable attention uses a fixed number of static reference points with corresponding offsets. Instead of traditional query and key projections, the LoRA adaptation adjusts these offsets and attention weights through low-rank matrices during finetuning.\\
Theoretically, the shifted reference points, also called sampling points, can be seen as queries, while the weights \(W_{attn}\) themselves can be seen as keys. Building upon this mental concept, LoRA is applied to the weights \(W_{offset}\) and \(W_V\), resulting in the decomposed matrices \(A_{offset}\), \(B_{offset}\), \(A_V\) and \(B_V\).\\
Mathematically, this LoRA adaptation of a deformable self-attention layer can be expressed as follows:
\begin{equation}
\label{eq:deformable-attention-lora}
    \begin{aligned}
        &\Delta p = x(W_{offset} + \frac{\alpha}{r}B_{offset}A_{offset})^T \\
        &\hat{A} = xW_{attn}^T \\
        &V = x(W_V + \frac{\alpha}{r}B_VA_V)^T \\
        &p = \hat{p} + \Delta p \\
        &A = softmax(\hat{A}) \\
        &Output = DeformableAttention(V, p, A)W_O^T
    \end{aligned}
\end{equation}
The deformable self-attention mechanism is visualized in Figure \ref{fig:lora-layers}b), where the trainable, decomposed LoRA weights are again colored red.
In this variant, the input sequence is used to compute the offsets, attention weights, and value projection.\\
Complementary, the cross-attention variant receives the value from a different sequence to facilitate the interaction between two sources. As in the self-attention variant, LoRA is applied to the offset and value projection weights. In the context of task adaptation, this updates the model's understanding of which points to look at by adjusting the offsets of the static reference points and how to interpret the inputs by adjusting the value projection.
%=================== Datasets and results ==========================%%
\section{Datasets and methodology}\label{sec:exp}
%========================== Datasets ==========================%
\subsection{Datasets}\label{sec:exp:subsec:datasets}
For a comprehensive evaluation across datasets of varying complexity, $4$ datasets were selected to serve as downstream tasks:
\begin{itemize}
  \item The Northumberland dolphin dataset $2020$ (NDD$20$) consists of $2,201$ images of both above and below water images, for a total of $4,402$ images and $6,102$ annotations of two different dolphin species \cite{Trotter2020NDD20AL}. As the second species is only present in the underwater images, no distinction between species was made during preparation, and the two types of images were concatenated and shuffled to create a training and test split. The images and annotations are of high quality, making the resulting dataset relatively less complex compared to the others.
  \item The ZeroWaste dataset is a unique and challenging benchmark dataset that poses numerous challenges, such as significant clutter, highly deformable and translucent objects, and a fine-grained difference between the object classes. It consists of real images taken from a conveyor belt in Materials Recovery Facilities (MRF). The fully annotated partition, \textit{ZeroWaste-f}, is unbalanced and consists of $4,503$ images with four possible classes: cardboard, soft plastic, rigid plastic, and metal. The SOTA for the AP is equal to $24.2$ \cite{Bashkirova2021ZeroWasteDT} and within this work, the dataset can be considered a medium complexity task.
  \item The Waste Inspection X-ray (WIXray) dataset introduced a real and novel problem of instance segmentation in X-ray images as a benchmark smart waste inspection. It consists of $5,038$ X-ray images (total of $30,881$ annotated waste items) with a constant resolution of $450$x$450$ and includes four general types and twelve categories of small objects: Recyclable (PlasticBottle, Can, Carton, GlassBottle, Stick, and Tableware), Residual (HeatingPad, Desiccant, and MealBox), Foodwaste (FoodWaste), and Hazardous (Battery and Bulb). The SOTA for the AP is equal to $46.85$ \cite{9880278} and due to the combination of rather small-sized images, occlusion, and overlapping instances due to the nature of X-ray scans, the dataset can be considered a high-complexity task within this research.
  \item The Cityscapes dataset \cite{Cordts_2016_CVPR} focuses on the semantic understanding of urban street scenes and consists of $5,000$ finely annotated and $20,000$ coarsely annotated images. It covers $30$ classes such as road, sidewalk, vehicle, and building, captured across $50$ cities over several months, during the daytime, and in good to medium weather conditions. The well-known benchmark dataset includes labels on pixel, instance, and panoptic segmentation levels. This study only focuses on the instance segmentation task with fine-grained labels, which consists of $5000$ images ($2,975$ for training, $500$ for validation, and $1,525$ for testing) containing annotations for $8$ classes. The current SOTA for the AP is equal to $49.3$, which was achieved by training the OpenSeeD model end-to-end \cite{zhang2023simple}.
\end{itemize}   
The four levels of complexity also result from their contextual divergence from the datasets used for pretraining the large pretrained models. It can be hypothesized that downstream tasks that are fundamentally different from the pretrained (common) knowledge require greater computational adjustments by the implemented adapter modules.
%========================== Ablation study ==========================%
\subsection{Experimental setup}\label{sec:ablation}
SEEM and Mask DINO are used as base models. The training procedure uses NVIDIA Quadro RTX $8000$ with a batch size of $4$ in all cases. The experiments were performed with different epochs for each dataset. Specifically, NDD$20$ was trained with $15$ epochs, Zerowaste with $10$ epochs, and both WIXray and Cityscapes with $20$ epochs. The performance of the models is evaluated by their best-resulting weights. The original optimizer (AdamW for both models) and loss function are kept throughout the study and after the sweep of different learning rate settings, the final base learning rates used were $0.001$ for adapter runs and $0.0001$ for traditional runs.
Notably, the Cityscapes dataset is trained with a more complex backbone (Swin-L) in the case of Mask DINO to allow a fair comparison to the latest SOTA results.\\
To highlight the parameter-efficiency, this approach is compared to several traditional finetuning settings without the use of LoRA and adapter modules. Since full finetuning requires updating $100\%$ of the parameters, which is computationally expensive, ablation studies on traditionally training explicit parts of the models were performed: 
Finetuning the full segmentation head (encoder + decoder), the decoder only, and class + mask embeddings only.\\
The different adapter settings include \{1, 2, 3, 4\} adapters per block ($I = 1, 2, 3, 4$)). The residual connection setting mentioned in section~\ref{sec:models-and-arch:subsec:adapters-arch:subsubsec:adapter-residual} is discussed further. Therefore, two additional experiments are performed on the NDD$20$ dataset (see Table~\ref{table1a} below). The setting \textbf{\bm{$\kappa$}}  refers to using only the attention key as the residual adapter input. In contrast, the setting \textbf{\bm{$\lambda$}}  refers to using the attention mechanism's output, and \textbf{\bm{$\kappa + \lambda$}} refers to adding both before feeding them into the adapter as a residual connection. The resulting AP values are documented in Table~\ref{table1a}.\\
A parameter sweep was conducted to identify the optimal scaling factor for the LoRA configurations. The configurations include \(r=\{2, 4, 8, 16\}\), while always using \(\alpha = r\) to attain a constant scaling factor of \(1\). This scaling was found to perform the best out of different \(\alpha\) values during a separate ablation study on the WIXray dataset, which can be found in Appendix \ref{sec:appendixA}.\\
SEEM integrates a dedicated language model that processes class names by injecting them into predefined prompts to generate the corresponding embeddings. Adapters and LoRA modules could potentially be introduced at this stage. However, as the primary function of the pretrained language model is to generate embeddings from textual input, regardless of the dataset, it was decided not to incorporate adapters or LoRA modules into the model at this stage, leaving this extension open for future work.
\def\arraystretch{1.2}
\begin{table}[htb!]
    \caption{Ablation study of the $4$ datasets with different finetuning settings and the number of parameters for each setting. Adapter and LoRA configurations also include the trainable class and mask embeddings.}
    \label{table1}
    \centering
    \begin{subtable}[t]{\textwidth}
    \caption{Ablation study for all four datasets. CE and ME refer to class and mask embedding, respectively. $\ast$ Uses Swin-L backbone}
        \label{table1a}
        \centering
        \resizebox{\textwidth}{!}{%
        \begin{tabular}{
            |p{0.2\textwidth}
            |>{\centering}p{0.1\textwidth}
            |>{\centering}p{0.1\textwidth}
            |>{\centering}p{0.1\textwidth}
            |>{\centering}p{0.1\textwidth}
            |>{\centering}p{0.1\textwidth}
            |>{\centering}p{0.1\textwidth}
            |>{\centering}p{0.1\textwidth}
            |>{\centering}p{0.1\textwidth}
            |>{\centering\arraybackslash}p{0.1\textwidth}|
        } \hline 

        \multirow{2}{*}{\emph{AP scores}} & \multicolumn{2}{c|}{\emph{NDD20}} & \multicolumn{2}{c|}{\emph{ZeroWaste}} & \multicolumn{2}{c|}{\emph{WIXray}} & \multicolumn{2}{c|}{\emph{Cityscapes}} & \multirow{2}{*}{\emph{Average}} \\

         & \emph{SEEM} & \emph{Mask DINO} & \emph{SEEM} & \emph{Mask DINO} & \emph{SEEM} & \emph{Mask DINO}  & \emph{SEEM} & \emph{Mask DINO}$\ast$ & \\ \hline

        Full Head & \textbf{79.2} & \textbf{78.98} & \textbf{25.97} & $24.94$ & \textbf{39.88} & \textbf{44.17} & \textbf{32.19} & $39.11$ & \textbf{45.55}
 \\ \hline 
        
        Only Decoder & $71.49$ & $66.28$ & $14.65$ & $14.26$ & $30.89$ & $28.72$ & $31.27$ & $36.83$ & $36.8$\\ \hline 
        
        Only CE \& ME & $67.79$ & $56.06$ & $5.17$& $6.99$ & $16.89$ & $6.62$ & $29.59$ & $34.16$ & $27.91$\\ \hline \hline
        
        $1$ Adapter ({$\kappa$}) & $76.26$ & $62.53$ & - & - & - & - & - & - & - \\ \hline 
        
        $1$ Adapter ({$\lambda$}) & $76.79$ & $75.55$ & - & - & - & - & - & - & - \\ \hline 
        
        $1$ Adapter ({$\kappa + \lambda$}) & $77.19$ & $76.54$ & $22.6$ & $22.66$ & $34$ & $36.96$ & $31.58$ & $37.23$ & $42.34$\\ \hline 
        
        $2$ Adapters ({$\kappa + \lambda$})& $77.72$ & $77.58$ & $24.2$ & $24.7$ & $34.45$ & $39.6$ & $31.7$ & $37.71$ & $43.461$\\ \hline
        
        $3$ Adapters ({$\kappa + \lambda$})& $77.67$ & $77.06$ & $22.9$ & $24.64$ & $35.28$ & $40.5$ & $32.08$ & $38.27$ & $43.55$ \\ \hline
        
        $4$ Adapters ({$\kappa + \lambda$})& $78.15$ & $76.71$ & $24.6$ & \textbf{25.02} & $35.27$ & $41.32$ & $32.04$ & $37.77$ & $43.86$\\ \hline \hline

        LoRA (r=2) & $77.29$ & $74.72$ & $22.6$ & $20.35$ & $32.97$ & $29.4$ & $31.22$ & $38.67$ & $40.9$ \\ \hline
        LoRA (r=4) & $77.06$ & $75.47$ & $23.1$ & $21.7$ & $34.95$ & $33.48$ & $31.26$ & $39.85$ & $ 42.11$ \\ \hline
        LoRA (r=8) & $76.89$ & $76.32$ & $23.9$ & $22.59$ & $35.06$ & $36.5$ & $31.77$ & $40.24$ & $42.91$ \\ \hline
        LoRA (r=16) & $76.52$ & $76.56$ & $24.5$ & $22.34$ & $34.01$ & $38$ & $31.3$ & \textbf{40.39} & $ 40.95$ \\ \hline

     \end{tabular}%
        }
        
    \end{subtable}
    
    \vspace{0.5cm} % Add space between the two tables

    \begin{subtable}[t]{\textwidth}
     \caption{Ablation study for the average number of trained parameters and their corresponding fraction of the resulting model.}
        \label{table1b}
        \centering
        \resizebox{\textwidth}{!}{%
        \begin{tabular}{
            |p{0.2\textwidth}
            |>{\centering}p{0.25\textwidth}
            |>{\centering}p{0.25\textwidth}
            |>{\centering\arraybackslash}p{0.25\textwidth}|
        }
            \hline
        \emph{\# Parameters} & \emph{SEEM} (Focal-L) & \emph{Mask DINO} (Resnet-$50$) & \emph{Mask DINO (Swin-L)} \\ \hline
        
        Full Head & $134.61M$ ($39.55\%$)& $28.56M$ ($54.91\%$) & $27.56M$ ($12.37\%$) \\ \hline 
        
        Only Decoder & $102.42M$ ($30.09\%$)& $14.51M$ ($27.89\%$) & $14.73M$ ($6.61\%$) \\ \hline 
        
        Only CE \& ME & $1.05M$ ($0.31\%$)& $0.33M$ ($0.63\%$) & $0.33M$ ($0.15\%$) \\ \hline \hline
        
        $1$ Adapter ({$\kappa + \lambda$}) & $4.21M$ ($1.23\%$)& $1.12M$ ($2.13\%$) & $1.12M$ ($0.50\%$) \\ \hline 
        
        $2$ Adapters ({$\kappa + \lambda$})& $7.37M$ ($2.13\%$)& $1.92M$ ($3.58\%$) & $1.92M$ ($0.86\%$) \\ \hline
        
        $3$ Adapters ({$\kappa + \lambda$})& $10.53M$ ($3.01\%$) & $2.71M$ ($4.99\%$) & $2.71M$ ($1.21\%$) \\ \hline
        
        $4$ Adapters ({$\kappa + \lambda$})& $13.7M$ ($3.88\%$)  & $3.51M$ ($6.35\%$) & $3.51M$ ($1.55\%$) \\ \hline \hline
        LoRA (r=2) & $1.15M$ ($0.34\%$) & $0.38M$ ($0.73\%$) & $0.38M$ ($0.17\%$) \\ \hline 
        LoRA (r=4) & $1.25M$ ($0.37\%$) & $0.43M$ ($0.83\%$) & $0.43M$ ($0.19\%$) \\ \hline 
        LoRA (r=8) & $1.44M$ ($0.42\%$) & $0.53M$ ($1.01\%$) & $0.54M$ ($0.24\%$) \\ \hline 
        LoRA (r=16) & $1.84M$ ($0.54\%$) & $0.73M$ ($1.39\%$) & $0.74M$ ($0.33\%$) \\ \hline 
    \end{tabular}%
        }
       
    \end{subtable}

\end{table}

In Table~\ref{table1}, CE represents the class embedding and ME represents the mask embedding. Unless otherwise noted, the residual \textbf{\bm{$\kappa + \lambda$}} setting is used for the rest of the paper.  
%========================== Discussion ==========================%
\section{Discussion}\label{sec:discussion}
This section presents a comparative analysis of adapter-based methods and LoRA, evaluating their performance and efficiency across different configurations and tasks. Quantitative and qualitative results, along with a comparison of inference speed are provided to assess their advantages and limitations.
%=========== Quantitative results ==========================%
\subsection{Quantitative results}
%=== DAnalysis of adapter performance ==========================%
\subsubsection{Analysis of adapter performance}
\label{sec:ada-discussion}
\begin{figure}[b!]
    \centering
    \includegraphics[width=0.9\linewidth]{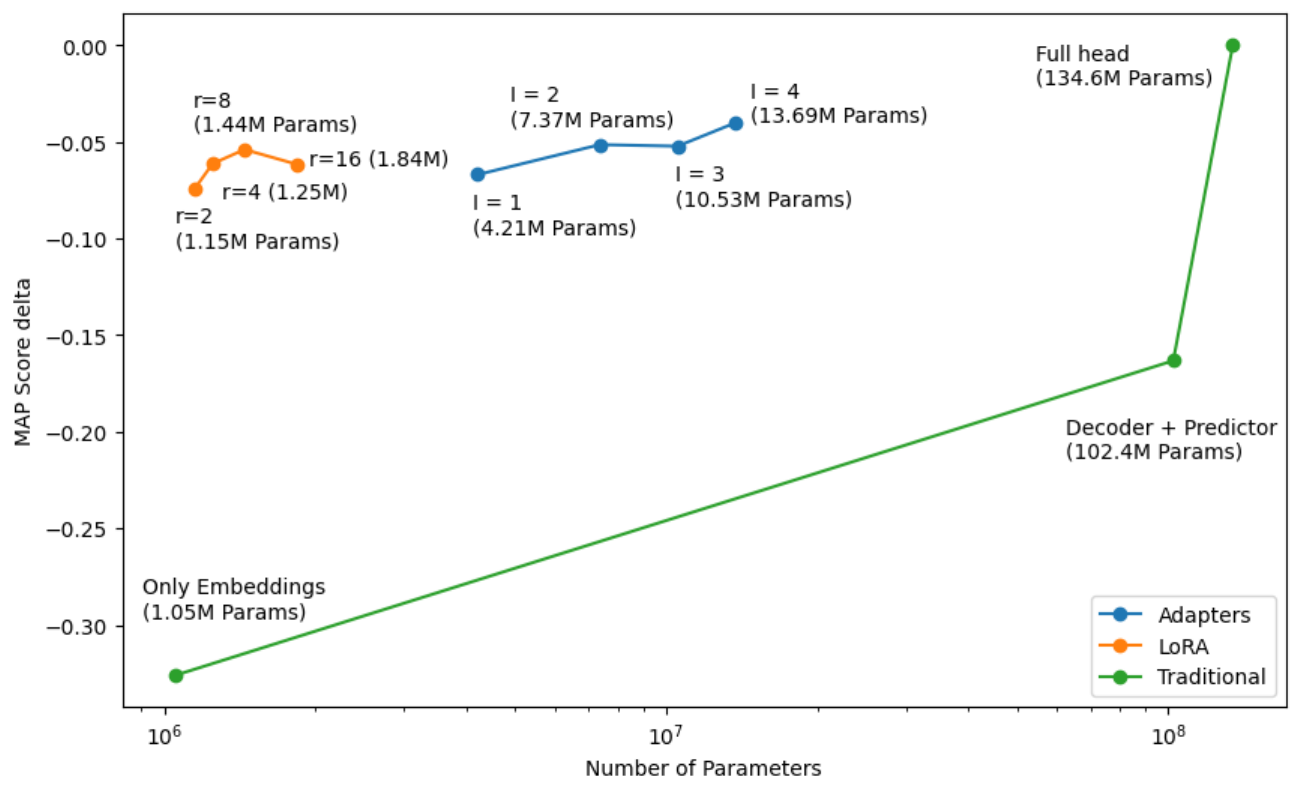}
    \caption{Delta (\(\delta\)) comparison of SEEM results}
    \label{fig:delta_comparison_seem}
\end{figure}

\begin{figure}[t!]
    \centering
    \includegraphics[width=0.9\linewidth]{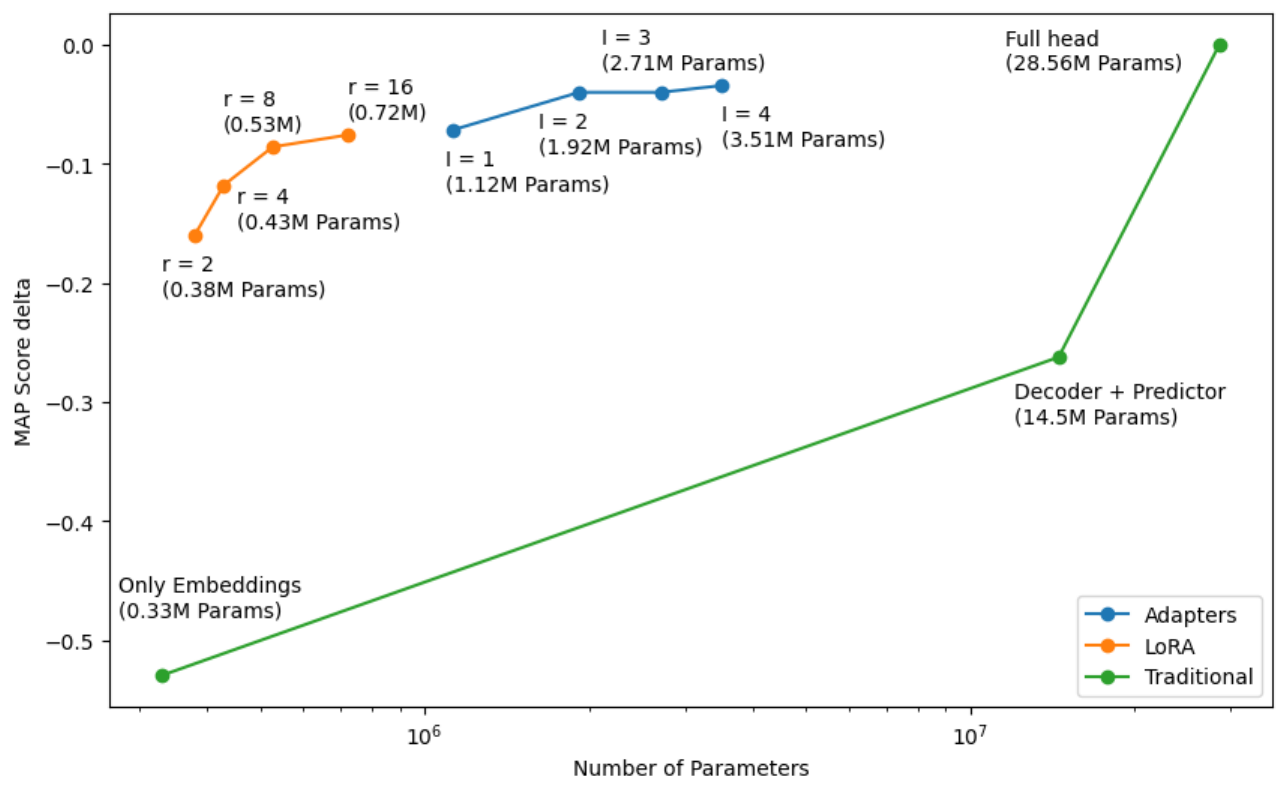}
    \caption{Delta (\(\delta\)) comparison of Mask DINO results (This does not include the Cityscapes results, because they were obtained with the larger Swin-L backbone).}
    \label{fig:delta_comparison_maskdino}
\end{figure}

This section provides an analysis of the performance of different adapter configurations, compared to full-head tuning in different datasets and settings.\\
For the AP, it was observed that full-head tuning generally gives slightly better results than adding adapters for both SEEM and Mask DINO in most configurations. However, there is an exception in the case of Mask DINO for the Zerowaste dataset, where the $4$ adapters setting outperforms both the full-head setting and even the SOTA. A more comprehensive comparison is visualized and discussed later using Figures~\ref{fig:delta_comparison_seem}, and~\ref{fig:delta_comparison_maskdino}.\\
In addition, it was found that tuning only the decoder or only the class and mask embeddings leads to a significant decrease in AP, emphasizing that removing tunable parameters in the final layers is inferior to adding adapters with similar or fewer parameters.\\
Furthermore, the performance of SEEM and Mask DINO varies across the datasets, but on average follows the trend of increasing performance as more adapters are added. Although the Mask DINO architecture with a ResNet-$50$ backbone is more than $6$ times smaller than SEEM, it could achieve comparable or better performance across the experiments. This suggests that the deformable attention-based detection branch, the contrastive denoising of the underlying DINO method, and the absence of a language encoder help the model to localize objects more efficiently before creating the segmentation masks. Mask DINO was also found to perform better in predicting small objects. Thus, a higher AP of Mask DINO regarding the WIXray dataset could be observed for this approach and full-head finetuning, as shown in Table~\ref{table1a}.\\
Table~\ref{table1b} shows a significant reduction in parameters when using adapter configurations compared to the full-head and only decoder settings. This reduction demonstrates the benefit of using adapters rather than modifying a subset of the underlying model to achieve parameter efficiency. Percentages are calculated as follows:
\begin{equation}
    percentage_{trained} = \frac{100 \times num_{trainable}}{num_{original} + num_{adapters}}
\end{equation}
For the Cityscapes benchmark, the best achieved AP score after $20$ epochs is equal to $38.27$ with the $3$ adapter setting and $39.11$ with the finetuning the full segmentation head of Mask DINO. Compared to the current SOTA for this benchmark dataset ($49.3$ AP for the multi-modal model "OpenSeeD" \cite{zhang2023simple}), the results indicate an increasing potential of the adapter finetuning, since the training was performed with a small number of additional parameters ($1.21\%$), rather than training the model end-to-end. The previous SOTA AP was equal to $46.7$, achieved by training the $372M$ parameter model "OneFormer" \cite{Jain_2023_CVPR} end-to-end for $90K$ optimization steps and a batch size of $16$, making up a total of $484$ epochs. To better compare the effectiveness of the adapters, the training of Mask DINO was extended with four adapters for an additional $30$ epochs, keeping all the hyperparameters constant. The best AP score obtained was $41.44$, while the model was still not observed to converge or overfit. Although the method still leaves a performance gap with SOTA for Cityscapes, it can be acknowledged that with further finetuning and optimization, the adapters could close this gap and provide an efficient framework for transferring pretrained knowledge to downstream tasks.\\
As a result, the adapter settings achieve competitive results with only $1.23 - 3.88\%$ (SEEM) and $0.5 - 6.35\%$ (Mask DINO) of the total model parameters. In contrast, finetuning the entire segmentation head results in a high percentage of trainable parameters, which is $39.55\%$ for SEEM and $54.91\%$ for Mask DINO.\\ 
To directly compare the methods with the full-head configuration, the average AP was calculated for each method across the $4$ datasets. Second, the relative difference between each of the resulting mean APs and the mean full-head AP is calculated as follows:
\begin{equation}
\delta_{m} = \frac{\overline{AP_m} - \overline{AP_{full head}}}{\overline{AP_{full head}}} \hfill,
\end{equation}
where $\overline{AP_m}$ is the mean for the APs for method $m$ and $\overline{AP_{full head}}$ is the mean of the APs for the full-head method.
In other words: The delta value (\(\delta\)) measures how far a method is away from the best-performing setting (full-head tuning) on a uniform scale. The results are shown on the y-axis in Figure~\ref{fig:delta_comparison_seem} for the SEEM model and Figure~\ref{fig:delta_comparison_maskdino} for the Mask DINO model. Additionally, the Figures show the number of trainable parameters on the x-axis.
Intuitively, the higher the number of adapters, the better the performance, and this proves to be the common ground for all four datasets. Therefore, finding the trade-off between the model precision and the number of parameters remains a challenge. The AP performance of using adapters is closely related to the finetuning of the full segmentation head. Although the traditional method yields slightly better results than the $2$ adapter setting, it requires significantly more parameters, specifically $18$ times more for SEEM and $15$ times more for Mask DINO (ResNet-$50$). Furthermore, adding $3$ or $4$ adapters has little impact on the performance while doubling the additive parameters. However, finetuning the decoder only shows a significant decrease in the average delta (\(\delta\)), accompanied by a considerable increase in the number of parameters for both models. Finetuning only the embedding weights does not capture enough information about the downstream task and results in an even larger reduction in the average delta \((\delta\)). This is exactly the problem where adapters and LoRA excel, as they are also placed in earlier layers and thus reduce the error accumulation throughout the network. \\
To summarize, the previous results indicate that $2$ to $3$ adapters per transformer block provide a fair and efficient transfer to instance segmentation tasks, offering the best trade-off between performance and the number of trained parameters among the tested settings. In addition to that, it is worth mentioning that none of the experiments showed a converging trend, leaving the potential for further improvement.
%==== Analysis of adapter performance ==========================%
\subsubsection{Analysis of LoRA performance}
\label{sec:lora-discussion}
This analysis aims to compare the performance of LoRA to full-head tuning and multiple adapter configurations based on the data provided in Table~\ref{table1}, and Figures \ref{fig:delta_comparison_seem}, \ref{fig:delta_comparison_maskdino}, and focuses on performance metrics and parameter efficiency.\\
In the NDD$20$ dataset, adapters show superior performance compared to LoRA, due to their higher number of parameters, which enables a more complex representation of water-specific visual characteristics. Moreover, the results in Table~\ref{table1a} show that increasing the rank of LoRA for SEEM results in a decline in performance, suggesting that these particular weight updates have a low intrinsic rank and higher-rank adaptations may negatively impact the model's ability to generalize effectively. Although LoRA outperforms $1$ adapter configurations within this dataset, it does not achieve the results seen with multiple adapter configurations in both models. For example, in SEEM, LoRA uses only $1.44M$ parameters compared to $4.21M$ for the $1$ adapter setup, allowing it to nearly match the performance of configurations that use almost \textit{three} times as many parameters. A similar analysis in Mask DINO shows that LoRA uses $0.53M$ parameters while the adapter configuration uses $1.12M$. These results suggest that while LoRA is highly efficient, the additional parameters in adapters may capture more precise settings matched to the NDD$20$ task.\\
In the ZeroWaste dataset, the multi-adapter configurations in the SEEM model consistently outperform LoRA, demonstrating a superior ability to handle the specific challenges of the dataset. These include high variability of deformable objects, fine-grained distinctions between overlapping waste categories, and complex backgrounds with frequent occlusions. The increased parameter capacity of the adapters proves crucial for learning robust representations of these highly variable waste objects. Even the $1$ adapter configuration shows competitive performance, outperforming LoRA in most cases, despite LoRA's parameter efficiency. This suggests that the task benefits significantly from the ability of the adapters to introduce task-specific layers. Similarly, in the Mask DINO model, all adapter configurations consistently outperform LoRA, suggesting that this architecture is particularly well suited to the adapter approach for the tasks presented in the ZeroWaste dataset. The superior performance of the adapters in both models demonstrates their effectiveness in capturing the complex features and relationships unique to waste classification tasks.\\
The WIXray dataset shows mixed results, emphasizing the varying effectiveness of finetuning methods based on model architecture. In the SEEM model, LoRA outperforms the $1$ adapter configuration and competes with multi-adapter setups despite using fewer parameters, indicating that LoRA's global updates can be effective for certain aspects of the WIXray task. However, in the Mask DINO model, LoRA consistently underperforms compared to all adapter configurations, similar to its performance in the ZeroWaste dataset. This suggests that Mask DINO may be less compatible with LoRA's update mechanism. The challenges of the WIXray dataset, such as the shift from natural images to X-ray scans, complicate the segmentation tasks. While LoRA shows some effectiveness in SEEM, adapters generally outperform it, especially in Mask DINO, due to their ability to introduce specialized processing essential for extracting material-specific features from small, overlapping objects in X-ray images. This context demonstrates the advantages of adapters, especially in architectures such as Mask DINO, while also illustrating the importance of the interaction between finetuning methods and model architectures in determining overall effectiveness for X-ray tasks.\\
The Cityscapes dataset illustrates remarkable results, especially for the Mask DINO model, where LoRA excels. In SEEM, LoRA's performance is similar to the other datasets, slightly outperforming the $1$ adapter setups but underperforming the multi-adapter setups. However, in Mask DINO, LoRA achieves the highest AP, outperforming all configurations, including full-head tuning, despite using only $0.54M$ parameters ($0.24\%$) compared to $27.56M$ ($12.37\%$) in full-head tuning. This success can be attributed to the fact that the Cityscapes dataset more closely matches common pretraining datasets that contain natural, visible-light images of everyday scenes. Consequently, LoRA's efficiency in updating pretrained weights is sufficient for optimal performance, as minor adjustments to existing features are sufficient in this context, while adapters may introduce over-complication and thus be less effective.\\
The parameter efficiency of LoRA makes it competitive in highly resource-constrained scenarios, as clearly shown in Figures \ref{fig:delta_comparison_seem} and \ref{fig:delta_comparison_maskdino}, while adapters tend to scale better with additional layers, showing different scaling characteristics. However, performance gains are not always linear with increasing parameters. For SEEM, increasing the trainable parameters for LoRA faces limitations at a rank of \(16\), indicating that the affected layers have a lower intrinsic rank compared to Mask DINO. As a result, increasing the rank beyond $8$ offers no significant benefit and even leads to a performance drop. From figures \ref{fig:delta_comparison_seem} and \ref{fig:delta_comparison_maskdino} it can be concluded, that the performance of LoRA quickly converges and further increasing the rank may not lead to better results, which matches the observations from the original LoRA publication in NLP \cite{Hulora}.
Besides these common limitations, the findings show that the effectiveness of finetuning methods varies depending on the task and model architecture.
%============ Qualitative results ==========================%
\subsection{Qualitative results}
\begin{figure}[h!]
    \centering
    \includegraphics[width=\linewidth]{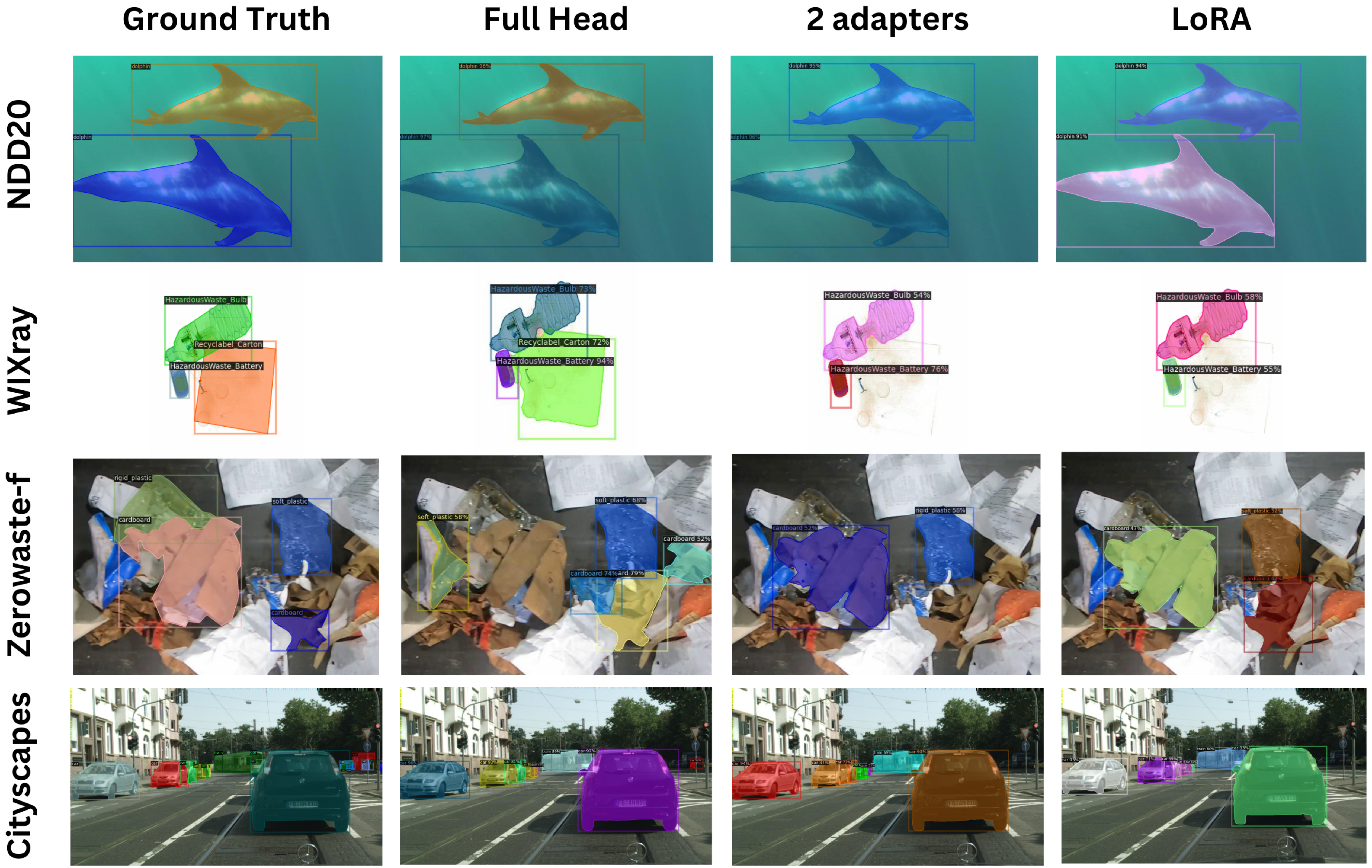}
    \caption{Four example-visualizations (cropped and scaled) per dataset for Mask DINO, showing the ground truth, full-head, $2$ adapter, and LoRA finetuning results.}
    \label{fig:example-preds}
\end{figure}

To provide more details on the visual representation of the results, one image of each dataset with the Mask DINO inference script is tested, which is provided by the authors, three times: With the best-finetuned weights from the full-head training, the $2$ adapter training and LoRA setting, which gives a reasonable parameter-performance trade-off across LoRA and the different adapter sizes. Figure~\ref{fig:example-preds} below visualizes the ground truth (GT) with the corresponding predictions of the $2$ inference runs. A confidence threshold of $0.5$ is implemented for the inference, effectively removing lower confidence predictions before visualization.\\
For the NDD$20$ dataset (first row), the predictions for all methods are highly accurate, almost achieving full intersection with the GT masks. The adapter method predicts the two masks with $1\%$ less confidence than the traditionally finetuned model. For the LoRA setting, the dolphin shapes are well-defined, and slightly less accurate, which results in a small difference in the exact contours compared to the GT and adapter methods.\\
In the WIXray dataset (second row), where the regions of interest mostly consist of bright contours on a white background. Although the adapters and LoRA successfully segment two out of three objects and correctly assign them to their respective classes, they have different levels of confidence, with LoRA showing the lowest confidence.\\
For the Zerowaste dataset (third row), the selected image contains four GT object annotations. Full-head method detects one additional object compared to GT, with higher confidence,  while the adapters and LoRA miss one object and have lower precision.\\
In the Cityscapes dataset (fourth row), the inference example follows the trend of almost exact object segmentation, regardless of visual distance, occlusion, and size of the instances. Again, the adapter and LoRA show lower confidence scores and also failed to detect the small car to the right compared to the full-head tuning. \\
The examples emphasize that employing the suggested adapter and LoRA finetuning method can indeed compete with the traditional finetuning approach in terms of pixel-wise object segmentation. However, further research is needed to compensate for the loss of trainable parameters fully and to increase the confidence of the model for individual detection.\\

\subsection{Inference speed}
The inference times for the tested datasets and configurations, image dimensions, and the respective numbers of classes are shown in Table \ref{table2} below, illustrating the impact of the PEFT methods on computational performance. The metric used is \(ms /iteration\), where each iteration is performed with a batch size of $1$.
The increase in inference times when using adapters can be attributed to the computations introduced by the additional layers. As the number of adapters increases, more parameters are added to the model, which slightly increases the computational requirements and the inference time. In contrast, LoRA generally imposes no computational overhead during inference, even with increasing rank. This is because the weights are re-composed and added to the base weights during initialization, which provides a clear advantage of LoRA in maintaining efficiency when finetuning large models.\\
For both models, the inference time varies across datasets. Specifically, three factors influence this variation: image resolution, the different numbers of classes, and the backbone used. Datasets with higher-resolution images require more processing time, as do datasets with a larger number of classes. While the open-set nature of SEEM ensures adaptability to different datasets, it does not reflect consistent processing times due to the variability in these characteristics.\\
In particular, the Cityscapes dataset shows a significant increase in inference time, especially when using Mask DINO. This is largely due to the Swin-L backbone used for Cityscapes, which is more computationally intensive than the ResNet$50$ backbone used for other datasets, and the high-resolution image size of $1024 \times 512$. The opposite behavior can be observed for the WIXray dataset, where the original images are of size $450 \times 450$.\\
On average, each increase of adapter repetition (\(I\)) was measured to introduce a delay of $1 - 2$ milliseconds per iteration. Again, this result indicates the trade-off between performance and computational complexity, achieving results between training the full segmentation head and LoRA. It's important to note that the inference speeds occur within the millisecond range, which may be negligible and unnoticeable during actual use.
\def\arraystretch{1.2}
\begin{table}[h!]
\caption{Image size, number of classes, and inference time (ms/iteration) of the model baseline, adapters, and LoRA across datasets.}
        \label{table2}
    \centering
        \resizebox{\textwidth}{!}{
        \begin{tabular}{
            |p{0.2\textwidth}
            |>{\centering}p{0.1\textwidth}
            |>{\centering}p{0.1\textwidth}
            |>{\centering}p{0.1\textwidth}
            |>{\centering}p{0.1\textwidth}
            |>{\centering}p{0.1\textwidth}
            |>{\centering}p{0.1\textwidth}
            |>{\centering}p{0.1\textwidth}
            |>{\centering\arraybackslash}p{0.1\textwidth}|
        } \hline 

        Dataset & \multicolumn{2}{c|}{\emph{NDD20}} & \multicolumn{2}{c|}{\emph{ZeroWaste}} & \multicolumn{2}{c|}{\emph{WIXray}} & \multicolumn{2}{c|}{\emph{Cityscapes}} \\ \hline

        Image Size & \multicolumn{2}{c|}{$512 \times 512$} & \multicolumn{2}{c|}{$512 \times 512$} & \multicolumn{2}{c|}{$450 \times 450$} & \multicolumn{2}{c|}{$1024 \times 512$} \\ \hline

        \# Classes & \multicolumn{2}{c|}{$1$} & \multicolumn{2}{c|}{$4$} & \multicolumn{2}{c|}{$12$} & \multicolumn{2}{c|}{$8$} \\ \hline

        Model & \emph{SEEM} & \emph{Mask DINO} & \emph{SEEM} & \emph{Mask DINO} & \emph{SEEM} & \emph{Mask DINO}  & \emph{SEEM} & \emph{Mask DINO} \\ \hline
        
        Model Baseline       & $65.43$ & $52.19$ & $85.06$ & $64.59$ & $54.5$ & $40.2$ & $135.69$ & $267.36$ \\ \hline
        
        $1$ Adapters ({$\kappa + \lambda$}) & $66.58$ & $54.92$ & $87.02$ & $66.7$ & $55.63$ & $42.89$ & $136.73$ & $271.25$ \\ \hline
        $2$ Adapters ({$\kappa + \lambda$}) & $67.52$ & $56.38$ & $88.16$ & $67.57$ & $56.91$ & $44.03$ & $137.18$ & $275.31$ \\ \hline
        $3$ Adapters ({$\kappa + \lambda$}) & $68.32$ & $58.29$ & $88.28$ & $69.07$ & $57.52$ & $46.08$ & $137.94$ & $283.92$ \\ \hline
        $4$ Adapters ({$\kappa + \lambda$}) & $70.36$ & $61.12$ & $88.79$ & $71.11$ & $58.52$ & $48.22$ & $138.43$ & $298.11$ \\ \hline
    
        LoRA (r=8)           & $65.76$ & $52.21$ & $85.28$ & $64.85$ & $54.11$ & $40.61$ & $135.86$ & $266.34$ \\ \hline
      
        \end{tabular}
        }
\end{table}
\section{Summary and conclusion}
\label{sec:comparison}
This study investigates the effectiveness of sequentially repeated adapters and LoRA for instance segmentation tasks in CV. Applying these techniques to SEEM and Mask DINO architectures and evaluating them on four different datasets (NDD20, ZeroWaste, WIXray, and Cityscapes) demonstrates their flexibility and scalability in transfer learning for large pretrained models.\\
Key findings and considerations include:
\begin{enumerate}
\item \textit{Trade-off between efficiency and performance}: While the study demonstrates significant parameter efficiency, it also reveals a consistent performance gap in certain scenarios. Adapter configurations used only $1.23-3.88\%$ (SEEM) and $0.5-6.35\%$ (Mask DINO) of the total model parameters, compared to $39.55\%$ (SEEM) and $54.91\%$ (Mask DINO) for full-head tuning. LoRA demonstrated even greater efficiency, using just $0.42\%$ of parameters for SEEM and about $1\%$ for Mask DINO with a ResNet-$50$ backbone. LoRA generally requires less computation and memory during training and inference due to its parameter-efficient design. The sequential alignment of adapters, while more parameter-intensive, can provide additional capacity beneficial for complex tasks or significant domain shifts. This trade-off between efficiency and performance raises important questions about the practical applicability of these methods in high-demand domains, where even small drops in performance can be critical. PEFT methods achieved comparable performance to traditional full-head tuning while significantly reducing the number of trainable parameters.
\item \textit{Optimal configurations}: Empirical evidence shows that $2$ to $3$ adapter repetitions provide the best trade-off between performance and parameter efficiency. Adding $4$ adapters tends to result in diminishing returns in terms of performance improvement while significantly increasing the number of additional parameters. In contrast, LoRA demonstrates high parameter efficiency, often outperforming $1$ adapter configurations despite using fewer parameters. 
\item \textit{Scalability}: Both PEFT methods showed performance improvements with an increase in trainable parameters. For Adapters, a more steady performance enhancement was observed compared to LoRA, potentially attributable to the linear nature of sequentially aligned adapters. Scaling the number of layers instead of the bottleneck dimension, or the rank in LoRA matrices, may better capture the underlying representations and dependencies of certain downstream tasks.
\item \textit{Dataset and architecture dependence}: The effectiveness of PEFT methods varied across datasets and model architectures, indicating that dataset complexity and architectural features play an important role in PEFT performance. For example, despite being six times smaller, Mask DINO achieved comparable or superior performance to SEEM in several experiments, likely due to its deformable attention-based detection branch and contrastive denoising approach. Specifically for the WIXray dataset, LoRA exhibited worse performance than adapters, suggesting that LoRA may be less effective on datasets where the images deviate significantly from the data used during pretraining.
\item \textit{Extended applicability}: The study successfully applied PEFT techniques to the multi-scale deformable attention module, extending their applicability beyond standard transformer architectures. This opens up possibilities for application to various SOTA methods based on DETR.
\end{enumerate}
These findings emphasize the importance of selecting PEFT methods based on model architecture and dataset characteristics. Furthermore, the empirical validation demonstrates the need to carefully consider the specific task requirements and computational constraints when choosing between LoRA and adapters.\\\\
There are several limitations to the study. First, there is some performance variability between tasks, with LoRA not performing as well on more complex datasets such as WIXray. This suggests that PEFT methods may not work equally well for every instance segmentation task. Another limitation is the increased inference time caused by adapters, which add extra layers and make them less suitable for real-time applications. While LoRA shows promise in vision tasks, its potential hasn't been fully explored beyond the specific use cases discussed, leaving many opportunities for broader application within vision models. In addition, by focusing on only two models, the study doesn't fully explore how well these PEFT methods would generalize to other architectures or tasks, such as object recognition or panoptic segmentation. Finally, there is a lack of detailed hyperparameter tuning, meaning that aspects such as learning rates and adapter placement were not thoroughly investigated, which could be key to further improving performance.\\\\
Future research could explore several directions to further optimize PEFT methods. A key area is to identify which tasks and model architectures are best suited to LoRA, potentially leading to guidelines on when it should be used via adapters. In addition, hybrid finetuning strategies combining adapters and LoRA, such as using adapters in higher layers and LoRA in attention layers, could be explored to optimize efficiency.
Although in this work LoRA was successfully applied to the deformable attention mechanism, several hyperparameters and configurations should be further investigated.
Researchers could also develop new variants of adapters, such as using convolutional layers or testing different injection points in the network, to improve performance and reduce early-stage errors. In addition, broader benchmarking on diverse sources like medical imaging, multi-modal learning and different vision tasks such as object detection or panoptic segmentation, could further test the adaptability and scalability of these methods. Overall, these efforts could pave the way for more effective and computationally efficient finetuning of large pretrained models.\\\\
%Future research directions include identifying the characteristics of tasks and model architectures that are most compatible with LoRA, which could lead to guidelines for when it should be preferred over adapter-based methods. Exploring different LoRA settings may also reveal better performance-efficiency trade-offs, as adjusting the rank based on the task or model and exploring the application of different weight matrices could optimize results. In addition, researchers could implement new adapter variants, such as using convolutional layers or other transformations, and explore different injection locations (e.g. backbone or post-convolutional layers) to improve task performance and reduce early-stage network errors. Finally, this study serves as a milestone for finetuning large pretrained vision models with fewer parameters and low computational complexity.\\
In conclusion, this study provides empirical evidence supporting the feasibility of PEFT methods for the efficient adaptation of large pretrained models in instance segmentation tasks. By demonstrating competitive performance with significantly reduced parameter counts, these techniques offer promising solutions for resource-constrained environments and fast adaptation scenarios. As the field progresses, these results contribute to ongoing efforts to balance computational efficiency and model performance in transfer learning for CV applications.
\vspace{6pt} 
%%%%%%%%%%%%%%%%%%%%%%%%%%%%%%%%%%%%%%%%%%
\newcommand{\david}{D.R.}
\newcommand{\nermeen}{N.A.B.}
\newcommand{\uwe}{U.H.}

\authorcontributions{Conceptualization, \nermeen and \david; methodology, \nermeen and \david; software, \david; validation, \nermeen and \david and \uwe; formal analysis, \nermeen and \david; investigation, \nermeen and \david; resources, \nermeen and \david; data curation, \david; writing---original draft preparation, \david; writing---review and editing, \david and \nermeen; visualization, \david; supervision, \uwe; project administration, \uwe; funding acquisition, \uwe
All authors have read and agreed to the published version of the manuscript.}
\funding{This work has been funded by the Ministry of Economy, Innovation, Digitization, and Energy of the State of North Rhine-Westphalia within the projects Digital.Zirkul\"ar.Ruhr and Circular Performer Emscher Lippe.}
\informedconsent{Not applicable.}
\dataavailability{{Full implementation of this work is published on Github:
\begin{itemize}
    \item \href{https://github.com/david-rohrschneider/SEEM-Adapter}{https://github.com/david-rohrschneider/SEEM-Adapter}
    \item \href{https://github.com/david-rohrschneider/MaskDINO-Adapter}{https://github.com/david-rohrschneider/MaskDINO-Adapter}
\end{itemize}
The final metrics and results are stored in Weights and Biases:
\begin{itemize}
    \item \href{https://api.wandb.ai/links/david-rohrschneider/ly1kdwzr}{https://api.wandb.ai/links/david-rohrschneider/ly1kdwzr} (SEEM)
    \item \href{https://api.wandb.ai/links/david-rohrschneider/rr18xj85}{https://api.wandb.ai/links/david-rohrschneider/rr18xj85} (Mask DINO).
\end{itemize}}}
\conflictsofinterest{The authors declare no conflicts of interest.} 
%%%%%%%%%%%%%%%%%%%%%%%%%%%%%%%%%%%%%%%%%%
\abbreviations{Abbreviations}{
The following abbreviations are used in this manuscript:\\

\noindent 
\begin{tabular}{@{}ll}
AP & Average Precision\\
CE & Class Embedding\\
CNN & Convolutional Neural Network\\
CV & Computer Vision\\
DETR & DEtection TRansformer\\
DINO & DETR with Improved Denoising Anchor Boxes\\
EMA & Exponentially Moving Average\\
FPN & Feature Pyramid Network\\
GT & Ground Truth\\
mIoU & mean Intersection-over-Union\\
LN & Layer Normalization\\
LoRA & Low Ranking Adaptation\\
ME & Mask Embedding\\
MLP & MultiLayer Perceptron\\
MRF & Materials Recovery Facilities\\
NLP & Natural Language Processing\\
PEFT & Parameter-Efficient Fine-Tuning\\
SAM  & Segment Anything Model\\
SEEM & Segment Everything Everywhere Model\\
SOTA & State-Of-The-Art\\
ViT & Vision Transformer\\
\end{tabular}
}

\appendixtitles{no} % Leave argument "no" if all appendix headings stay EMPTY (then no dot is printed after "Appendix A"). If the appendix sections contain a heading then change the argument to "yes".
\appendixstart
\appendix
\section[\appendixname~\thesection]{}\label{sec:appendixA}
The appendix presents the Table \ref{tab:alpha_r} that evaluates four \(\alpha\) values with constant \(r = 8\). This comparison provides insights into the model performance under different scaling conditions for WIXray dataset.
\begin{table}[h!]
\caption{The evaluation of different scaling configurations \(\alpha\) for LoRA.}
\label{tab:alpha_r}
\centering
\begin{tabular}{|c|c|c|c|c|}
\hline
WIXray dataset & \(\alpha = 1\)  & \(\alpha = 2\)  & \(\alpha = 4\)   & \(\alpha = 8\)  \\ \hline
   AP   & $32.2$   &  $33.95$   & $34.86$    & $36.5$   \\ \hline
\end{tabular}
\end{table}
\begin{adjustwidth}{-\extralength}{0cm}

\reftitle{References}

\PublishersNote{}
\end{adjustwidth}
\end{document}